\documentclass[10pt,twocolumn,letterpaper]{article}

\usepackage[pagenumbers]{iccv} 

\usepackage{graphicx}
\usepackage{amsmath}
\usepackage{amssymb}
\usepackage{booktabs}
\usepackage{blindtext}

\usepackage{times}
\usepackage{epsfig}
\usepackage{enumitem}

\usepackage{multirow}
\usepackage{amssymb}
\usepackage{algorithm}
\usepackage{algorithmic}
\usepackage{bm}

\definecolor{iccvblue}{rgb}{0.21,0.49,0.74}
\definecolor{slategray}{RGB}{47, 79, 79}
\definecolor{royalblue}{RGB}{65, 105, 225}
\definecolor{darkgreen}{RGB}{0, 100, 0}

\newcommand{\CommentColorDarkGreen}[1]{\textcolor{darkgreen}{#1}}
\usepackage[pagebackref,breaklinks,colorlinks,allcolors=iccvblue]{hyperref}

\def\paperID{9897} 
\def\confName{ICCV}
\def\confYear{2025}

\title{Diffusion-based Data Augmentation and Knowledge Distillation with Generated Soft Labels Solving Data Scarcity Problems of SAR Oil Spill Segmentation}

\author{%
Jaeho Moon\textsuperscript{1}\footnotemark[1] \quad\quad Jeonghwan Yun\textsuperscript{1}\footnotemark[1] \quad\quad Jaehyun Kim\textsuperscript{1}\footnotemark[1] \quad\quad Jaehyup Lee\textsuperscript{2}\footnotemark[2] \quad\quad Munchurl Kim\textsuperscript{1}\footnotemark[2]\\ [5pt]
\textsuperscript{1}Korea Advanced Institute of Science and Technology  \\[2pt]
\textsuperscript{2}Kyungpook National University \\
\small{\url{https://kaist-viclab.github.io/DAKTer-site/}}
}

\begin{document}
\maketitle
\begin{abstract}
Oil spills pose severe environmental risks, making early detection crucial for effective response and mitigation.
As Synthetic Aperture Radar (SAR) images operate under all-weather conditions, SAR-based oil spill segmentation enables fast and robust monitoring.
However, when using deep learning models, SAR oil spill segmentation often struggles in training due to the scarcity of labeled data.
To address this limitation, we propose a diffusion-based data augmentation with knowledge transfer (DAKTer) strategy. 
Our DAKTer strategy enables a diffusion model to generate SAR oil spill images along with soft label pairs, which offer richer class probability distributions than segmentation masks (i.e. hard labels).
Also, for reliable joint generation of high-quality SAR images and well-aligned soft labels, we introduce an SNR-based balancing factor aligning the noise corruption process of both modalilties in diffusion models.
By leveraging the generated SAR images and soft labels, a student segmentation model can learn robust feature representations without teacher models trained for the same task, improving its ability to segment oil spill regions.
Extensive experiments demonstrate that our DAKTer strategy effectively transfers the knowledge of per-pixel class probabilities to the student segmentation model to distinguish the oil spill regions from other look-alike regions in the SAR images. 
Our DAKTer strategy boosts various segmentation models to achieve superior performance with large margins compared to other generative data augmentation methods.
\end{abstract}

{
  \renewcommand{\thefootnote}%
    {\fnsymbol{footnote}}
  \footnotetext[1]{Co-first authors, equal contribution;}
  \footnotetext[2]{Co-corresponding authors.}
}
\section{Introduction}
\label{sec:intro}

\begin{figure}
    \centering
    \includegraphics[width=\columnwidth]{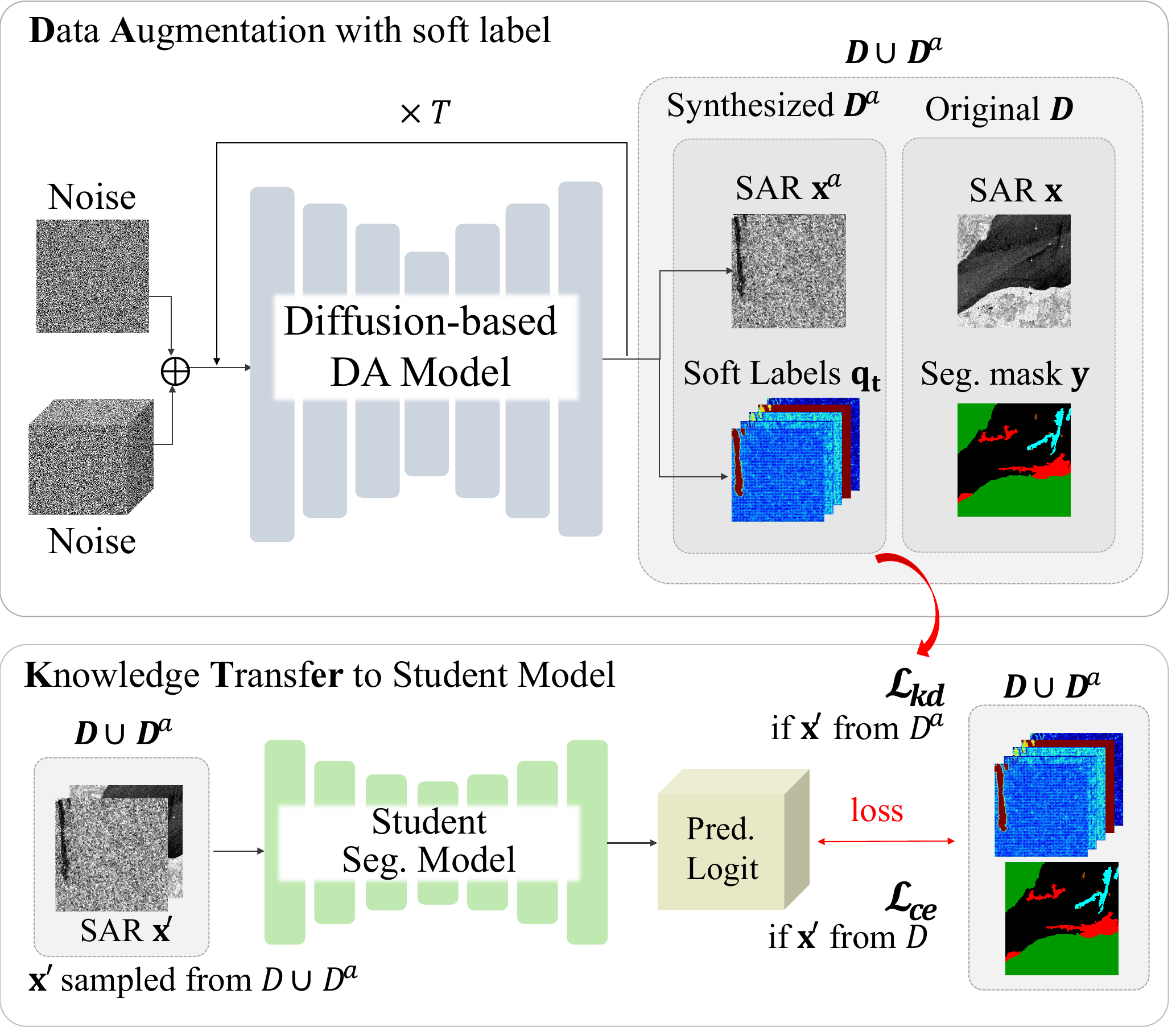}
     \caption{Overview of our Data Augmentation and Knowledge Transfer (DAKTer) strategy. A diffusion-based data augmentation model synthesizes an augmented dataset $\mathcal{D}^a$ consisting of image-soft label pairs. Additionally leveraging $\mathcal{D}^a$, a student segmentation model can be trained with more informative soft labels, compared to segmentation masks (hard labels).}
    \label{fig:figure_1}  
    \vspace{-5mm}
\end{figure}

Oil spills, caused by offshore drilling, shipping accidents, and natural seepage, release vast amounts of oil into the ocean, posing severe, often irreversible threats to marine ecosystems and coastal economies. 
These spills can lead to long-term environmental degradation, harming marine life, and disrupting local industries such as fisheries and tourism.
To mitigate these impacts, early detection and accurate segmentation of oil spill regions using remote sensing data are essential.
Among remote sensing modalities, Synthetic Aperture Radar (SAR) stands out due to its ability to acquire data in all weather and lighting conditions by emitting microwave signals and measuring the reflected backscatter from the Earth’s surface \cite{intro_speckle_noise_jh, intro1_SARimage_description, intro2_SARimage_description, intro9_SARimage_description}. 
Also, SAR imagery is particularly useful for oil spill segmentation, as oil reduces surface roughness, causing the affected areas to appear as dark spots \cite{intro3_darkspot, intro4_darkspot}. 
Recent advances in deep learning-based SAR oil spill segmentation enable faster response and effective decision-making \cite{intro5_oilspill, intro6_oilspill, intro7_oilspill_lookalike}.

However, SAR-based oil spill segmentation faces a significant challenge: the scarcity of labeled data. 
This is because real oil spill events are infrequent, making SAR images capturing such incidents limited. 
Furthermore, obtaining accurate annotations is complex due to the inherent speckle noise in SAR imagery \cite{intro15_specklenoise,intro10_speckle_noise}, which can obscure oil spill boundaries. 
The scarcity of high-quality labeled data hinders the performance of deep learning-based segmentation models, which require diverse samples to learn robust features. 
To tackle this problem in other semantic segmentation domains, recent studies \cite{intro20_da_label_anatomic_ctr_med_seg, intro19_da_label_nuclei_image_seg, wu2023diffumask, satsynth} introduce to generate image-label pairs utilizing diffusion models \cite{ddpm} for data augmentation (DA).

In the SAR oil spill segmentation domain, we observed two key aspects during the application of current diffusion-based DA approaches.
First of all, we observed that diffusion models often struggle to jointly generate SAR images and segmentation labels.
This problem comes from the different noise sensitivity between the two modalities during the noise corruption process of diffusion models, necessitating a balancing mechanism to ensure stable joint generation.
Second, current diffusion-based DA methods synthesize only image and hard-label pairs for downstream segmentation tasks.
However, we observed that diffusion models can produce soft labels that contain relative class probabilities reflecting the semantic understanding ability of diffusion models \cite{wu2023diffumask, diffuse_attend_segment_zeroshot_seg_diff, label-efficient_seg_diff_mlp}.
Unlike hard labels, soft labels provide probability distributions over multiple classes per pixel, capturing semantic similarities.

Based on these observations, we propose \textbf{a diffusion-based Data Augmentation and Knowledge Transfer (DAKTer) strategy} (i) to address the scarcity of labeled SAR oil spill data and (ii) to utilize augmented (generated) data for effective training of segmentation models with generated soft labels.
In our DAKTer strategy, we train DDPM \cite{ddpm} as a diffusion-based DA model to jointly synthesize both realistic SAR oil spill images and their soft labels that contain probability distributions over multiple classes per pixel.
By leveraging both generated SAR images and soft labels, we transfer the knowledge contained in the soft labels from DDPM into the student segmentation model learning process of segmenting oil spill regions. 
Also, to ensure stable joint generation, we propose an SNR-based balancing factor that aligns noise sensitivity across two modalities during the noise corruption process in diffusion models by leveraging the signal-to-noise ratio (SNR).
Our comprehensive experiments demonstrate that our DAKTer effectively mitigates the data scarcity problem in SAR oil spill segmentation, and also significantly boosts the performance of various segmentation models \cite{exp_deeplabv3+, cbdnet, cheng2021mask2former, experiment_segformer} on the OSD \cite{related_oilspill_compare_all}, SOS-ALOS \cite{cbdnet}, and SOS-Sentinel \cite{cbdnet} datasets. 
Specifically, our DAKTer enhances the segmentation performance of the state-of-the-art segmentation model, SegFormer \cite{experiment_segformer}, increasing $3.28\%$ points from $67.46\%$ to $70.74\%$ in terms of mIoU on the OSD dataset \cite{related_oilspill_compare_all}, compared to the model trained with the only original dataset \cite{related_oilspill_compare_all}.

\cref{fig:figure_1} illustrates our DAKTer strategy. 
A diffusion-based DA model is trained to generate SAR oil spill images and their corresponding soft labels using the original training dataset $\mathcal{D}$ that contains a limited amount of real SAR oil spill images and their per-pixel hard (one-hot) labels.
After training, the diffusion-based DA model synthesizes an augmented dataset $\mathcal{D}^a$ including SAR oil spill images and their soft labels. 
Leveraging both $\mathcal{D}$ and $\mathcal{D}^a$, we train a student segmentation model with knowledge distillation using the generated soft labels, boosting their segmentation performance with large margins.
Our contributions are summarized as follows:

\begin{itemize}
    \item We propose a diffusion-based Data Augmentation and Knowledge Transfer (DAKTer) strategy to overcome data scarcity problems in SAR oil spill segmentation.
    \item We introduce an SNR-based balancing factor that enables the modality-balanced joint generation of SAR oil spill images and their soft labels using diffusion models.
    \item In our DAKTer strategy, the generated realistic SAR oil spill images and their corresponding soft labels can be effectively used to train downstream student segmentation models by transferring the knowledge obtained from the diffusion-based DA model.
    \item Our proposed strategy \textit{significantly} boosts the performance of various student segmentation models on the public SAR oil spill datasets \cite{related_oilspill_compare_all, cbdnet}.
\end{itemize}

\section{Related Work}
\label{sec:related_work}

\subsection{Semantic Segmentation on SAR images}
  
Deep learning-based semantic segmentation is widely applied in remote sensing, utilizing Electro-Optical and SAR imagery \cite{related_remotesegmentation1, related_remotesegmentation2, related_remotesegmentation3}. 
For oil spill segmentation in SAR images, various deep learning models have been employed, including U-Net \cite{related_oilspill_unet, related_oilspill_compare_all,related_oilspill_quad}, LinkNet \cite{ related_oilspill_compare_all}, DeepLab \cite{related_oilspill_deeplab,related_oilspill_compare_all}, Mask R-CNN \cite{related_oilspill_maskrcnn} and Dual Attention \cite{related_oilspill_dualattention}. 
Recently, SAM-OIL \cite{SAM_OIL} proposed a two-stage approach, predicting bounding boxes for oil spill using YOLOv8 \cite{mmyolo2022} and performing segmentation based on the boxes using SAM \cite{related_sam}.

For training deep learning models to perform oil spill segmentation based on SAR images, Zhu \etal \cite{cbdnet} introduced the SAR Oil Spill (SOS) dataset, comprising 8,070 labeled SAR images from ALOS and Sentinel satellites, and Krestenitis \etal \cite{related_oilspill_compare_all} proposed the Oil Spill Detection (OSD) dataset, consisting of 1,112 annotated SAR images.
While deep learning models have achieved promising results on these datasets, data scarcity still remains a major limitation. 
This challenge arises not only due to the event-specific nature of oil spills but also from the difficulty of generating accurate ground truth masks for oil spill and other regions in SAR imagery. 
Speckle noise \cite{intro15_specklenoise,intro10_speckle_noise} and environmental variations make precise annotation both labor-intensive and error-prone, further restricting model performance.
To overcome the data scarcity problems, we propose our DAKTer strategy that leverages the generative power of diffusion models to jointly synthesize realistic SAR images and their soft labels, leading to substantial performance improvements in SAR oil spill segmentation.

\subsection{Data Augmentation} 

Recent advancements in generative models, particularly GANs \cite{goodfellow2020gan} and diffusion models \cite{ddpm, rombach2022high}, have enabled the generation of diverse images with corresponding label pairs for data augmentation (DA).
While previous approaches \cite{effectivedataaugmentationdiffusion, is_synth_data_ready_for_KD, da_label_clss_dreamda, intro21_da_wo_syndataimprove_clss,GEODIFFUSION, intro16_DA_label_object_ctrldiff} generate images conditioned on predefined labels (e.g., class names or bounding boxes), they fail to increase label diversity.
To overcome this, other approaches \cite{semgan, diffusionengine, ScribbleGen, intro19_da_label_nuclei_image_seg, satsynth, jointnet,wu2023diffumask} jointly generate image-label pairs to ensure the diversity in both.
For semantic segmentation, Yu \etal \cite{intro19_da_label_nuclei_image_seg} generates images first, and then generates masks conditioned on the generated images.
On the other hand, SemGAN \cite{semgan} generates images and segmentation mask pairs simultaneously.
As the image generation diffusion model has shown to have knowledge for semantic segmentation in \cite{diffuse_attend_segment_zeroshot_seg_diff, label-efficient_seg_diff_mlp}, Wu \etal \cite{wu2023diffumask} extracts segmentation masks from generated images using text-guided cross-attention information as a knowledge from diffusion. 
For satellite image segmentation tasks, SatSynth \cite{satsynth} introduces a diffusion-based DA method for generating pairs of colored satellite images with one-hot segmentation masks.

However, for SAR oil spill segmentation tasks, traditional DA techniques such as flipping, rotation, and scaling remain widely used but fail to increase semantic diversity \cite{related_oilspill_unet, related_oilspill_dualattention}.
In this paper, we handle the data scarcity problem in SAR oil spill segmentation, by jointly synthesizing pairs of SAR oil spill images and their per-pixel soft labels which preserve richer class probabilities rather than hard labels that recent DA methods used to utilize \cite{semgan, satsynth}.
This allows us to leverage the diffusion model’s knowledge of semantic understanding while demonstrating the effectiveness of soft labels for SAR oil spill segmentation.

\subsection{Knowledge Distillation} 

Knowledge distillation (KD) refers to the process of conveying information from a teacher model to a student model. This concept was first introduced by Hinton \etal \cite{hinton2015distillingknowledgeneuralnetwork}, where a large or more complex deep neural network (teacher) transfers softened output predictions, or soft labels, to a smaller or simpler network (student) to improve its performance. 
Following this, various methods have been proposed for transferring knowledge using output logits  \cite{Zhao_2022_CVPR, Zhang_2019_CVPR, what_KD, SKD, li2021online} and intermediate features \cite{wang2020intra,liu2019structured, fitnet, sp, AT, CAT, reviewkd, park2019relational}. 
For semantic segmentation tasks, \cite{SKD} utilizes the class probabilities derived from output logits to transfer knowledge from teacher models to student models. 
These logits not only encode predicted class probabilities, but also capture uncertainty, semantic context, and relationships between similar classes, helping the student learn smoother and more consistent boundaries, especially for visually similar categories.
Even when the teacher and student models are designed for different tasks, the knowledge transfer through logits has been shown to be effective \cite{multi-modal_kd1,multi-modal_kd2,multi-modal_kd3,multi-modal_kd4}.

KD has also been studied to address data dependency during the training of neural networks.
For inaccessible training data, the data-free KD approaches \cite{Chen_2019_ICCV, kd:up_to_100x, ye2020data-free-kd} enable knowledge transfer from teacher to student models by training both on the datasets generated by GAN-based models \cite{goodfellow2020gan}.
Unlike the data-free KD methods, our DAKTer strategy focuses on handling the problem of scarce labeled training data.
Moreover, our DAKTer strategy directly transfers knowledge from synthesized soft labels in generated datasets using diffusion models \cite{ddpm}, without training any teacher networks that perform the same tasks.

\section{Method}
\label{sec:method}

\subsection{Motivation}

\begin{figure}
    \centering
    \includegraphics[width=0.7\columnwidth]{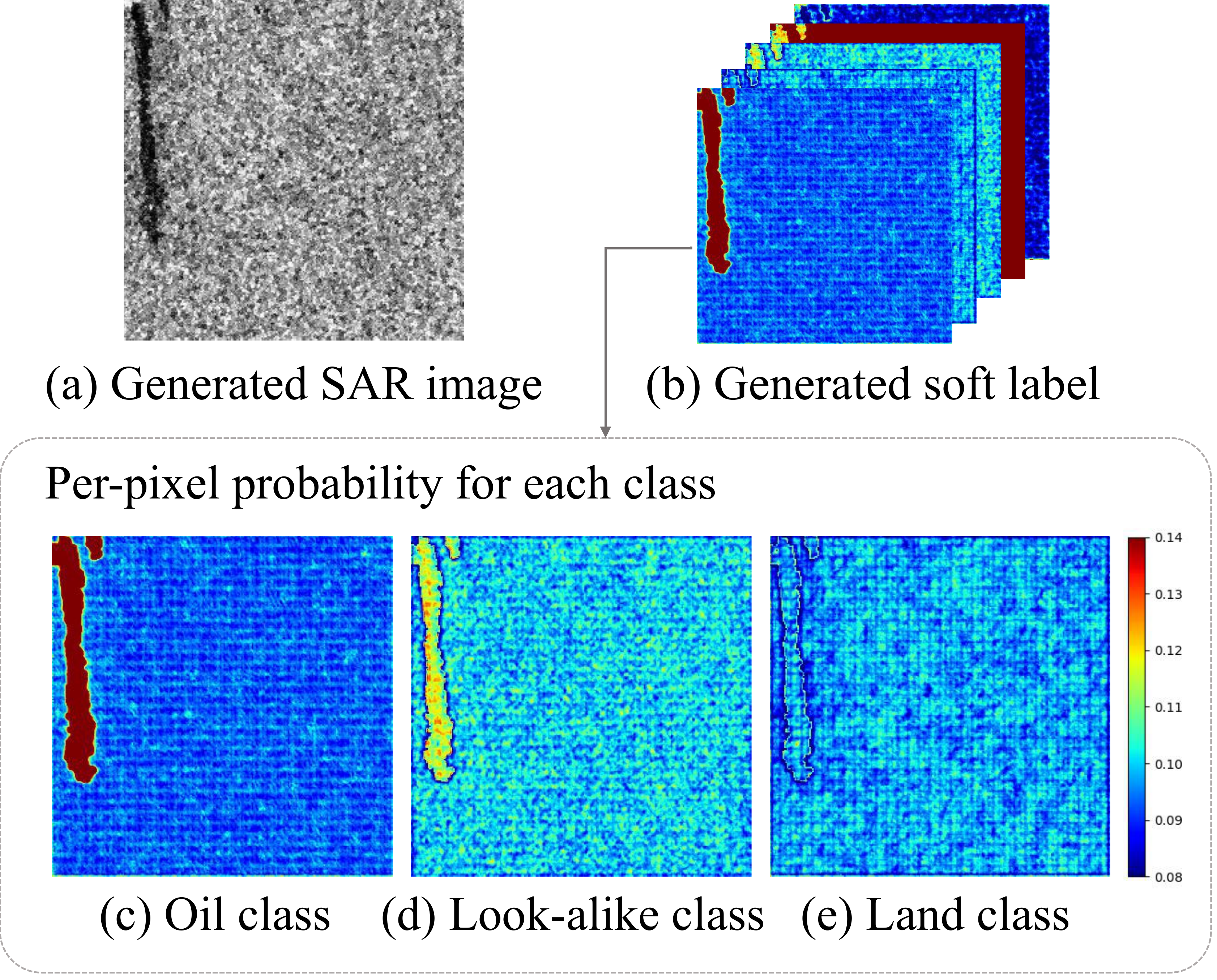}
     \caption{Visualization of generated outputs from DDPM \cite{ddpm} trained with our DAKTer strategy. (a) Generated SAR image with oil spill region (the dark region on the left). (b) Generated soft label containing probability (prob.) maps for each class (scaled for better visualization). (c) Prob. values are high in the oil spill region. (d) As `look-alike' class is shown to be dark in SAR images similar to `oil-spill' class, the prob. values are relatively higher (yellow regions) in the oil spill region than those in other regions. (e) As `Land' class appears as bright in SAR images (distinctive to 'oil-spill' class), the prob. values are low in the whole region.
     }
    \label{fig:motivation}  
    \vspace{-2mm}
\end{figure}

Data scarcity, which is especially a critical issue for SAR oil spill segmentation, induces limited segmentation performance of deep learning networks.
To mitigate this issue, previous work \cite{semgan, satsynth} suggested data augmentation techniques to synthesize labeled datasets for segmentation tasks based on generative models, such as GANs \cite{goodfellow2020gan} and diffusion models \cite{ddpm, rombach2022high}.
A key advantage of diffusion models over other generative approaches is their strong semantic understanding, which enables them to capture complex spatial relationships within images \cite{wu2023diffumask} and effectively perform zero-shot semantic segmentation \cite{label-efficient_seg_diff_mlp, diffuse_attend_segment_zeroshot_seg_diff}.

In the context of the joint generation of SAR oil spill images and segmentation masks, semantic knowledge of diffusion models allows them to produce semantically meaningful values in the labels over generated SAR oil spill images. 
We observed that the generated soft labels preserve the relative probability distribution among classes.
Specifically, the pixels belonging to similar classes tend to correspond to high probabilities, reinforcing their class relationships.
On the other hand, the pixels corresponding to dissimilar classes retain low probabilities against the similar classes, effectively containing inter-class distinctions.
\cref{fig:motivation} shows the per-pixel probability map of `oil-spill', `look-alike', and `land' classes in a synthesized soft label.
In SAR images, pixels belonging to the `oil-spill' and `look-alike' classes are shown to be dark while pixels for `land' are bright and have unique structures (as shown in \cref{fig:OSD_generation_qual}).
As shown in the probability values of the `look-alike' class, the probability values are high in oil spill regions (yellow regions), demonstrating that synthesized soft labels tend to have a semantically aligned probability distribution.
Conversely, the probability values for `land' class are consistently low for the whole region.
This suggests that diffusion-based generated soft labels can provide more informative supervision than hard labels to train student segmentation networks.
Motivated by this insight, we propose a diffusion-based data augmentation and knowledge transfer (DAKTer) strategy that enhances the performance of the segmentation models through supervision with soft labels.

\subsection{Method Overview}

Our proposed diffusion-based DAKTer strategy has three stages for data augmentation and knowledge transfer.

\noindent\textbf{Learning joint distributions on pairs of SAR oil spill images and soft labels.}
The first stage involves training DDPM \cite{ddpm} to learn the joint distribution of SAR images and segmentation maps from the given original dataset \(\mathcal{D}\) as shown in \cref{fig:figure_1}.
To ensure a balanced noise corruption process of diffusion models between the two modalities of SAR images and segmentation masks, we propose an SNR-based balancing factor.
Additionally, we use a cross-entropy loss to generate soft labels containing semantically aligned probabilistic values for their corresponding classes.

\noindent\textbf{Synthesizing realistic SAR oil spill images and soft labels.}
Once trained, the DDPM acts as a data generator, synthesizing SAR oil spill images along with their soft labels, forming an augmented dataset denoted as \(\mathcal{D}^a\).
Our approach generates entirely new image-label pairs, overcoming the data scarcity problem with large diversity.

\noindent\textbf{Knowledge transfer using soft labels for supervision of student segmentation networks.} 
In the final stage, we train student segmentation models using a combination of \(\mathcal{D}\) and \(\mathcal{D}^a\).
Since \(\mathcal{D}^a\) contains logit-based soft labels, we train student models using a knowledge distillation loss \cite{hinton2015distillingknowledgeneuralnetwork}, which enables them to learn from supervision with soft labels provided by the diffusion model.
This improves segmentation performance by giving richer information compared to relying only on hard labels.

\subsection{Learning Joint Distribution of SAR Oil Spill Images and Soft Labels}
\label{sec:learning_joint_distribution}

\begin{figure}
    \centering
    \includegraphics[width=\linewidth]{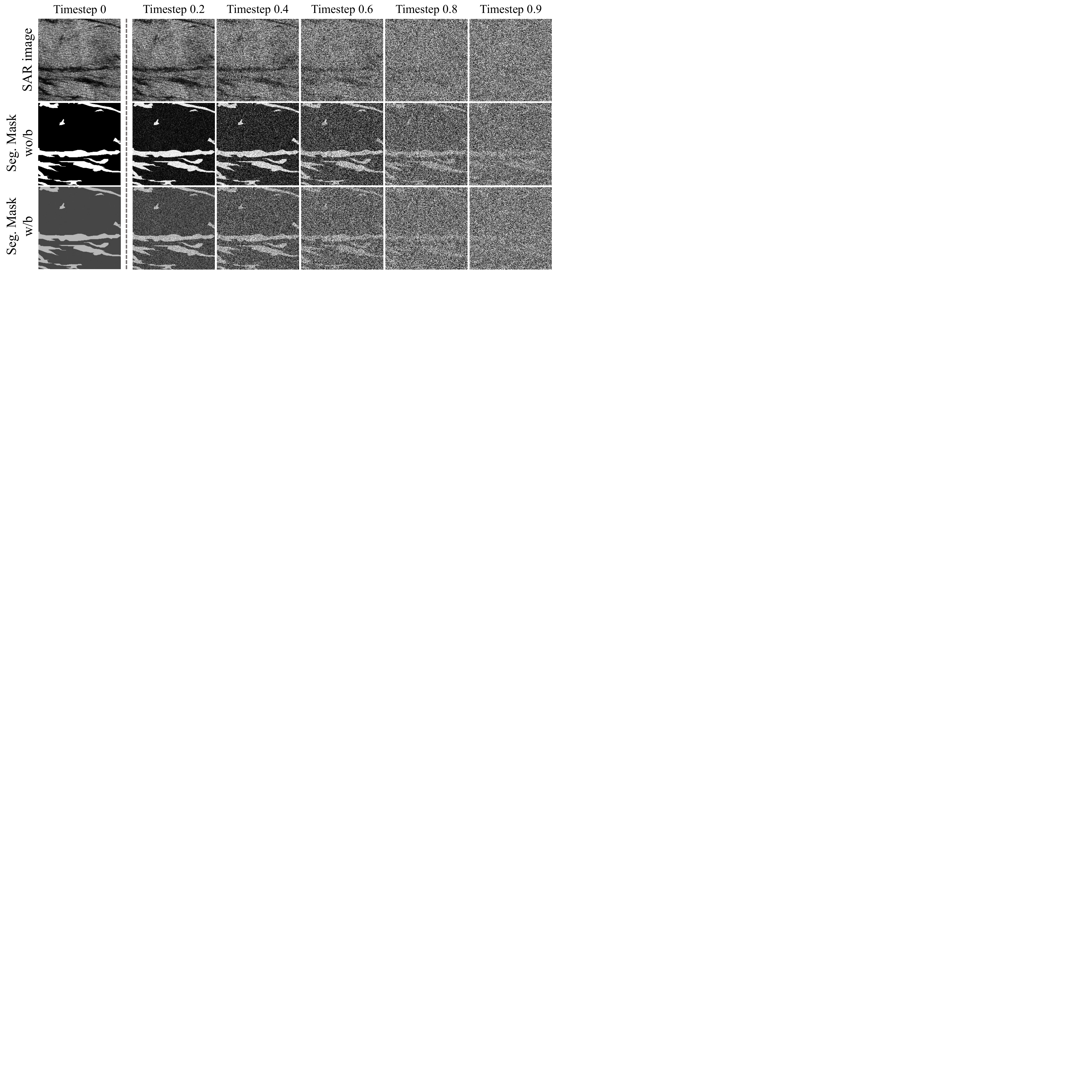}
    \caption{Effect of our SNR-based balancing factor \(b\) on the noise corruption process  in DDPM \cite{ddpm}. Without applying \(b\), segmentation masks always have more structural information compared to SAR images at the same timestep. With applying \(b\), noise corruption is balanced across two modalities, preserving comparable information during joint generation.}
    \label{fig:balancing_factor}
\end{figure}

\subsubsection{Balancing Joint Generation}
\label{sec:balancing_joint_generataion}

The conventional training strategy of DDPM is to reconstruct clean images from noise-corrupted inputs.
We extend this to learn to generate clean image-label pairs from their noise-corrupted counterparts, thereby modeling their joint distribution. Given a SAR image $\mathbf{x}_0  \in \mathbb{R}^{H \times W\times 1}$ and its corresponding normalized segmentation mask $\mathbf{y}_0 \in \{-1, 1\}^{H \times W \times C} $ where \(C\) is the number of classes, the corrupted image-label pair \(\{\mathbf{x}_t, \mathbf{y}_t\}\) at a specific timestep $t \in [0, T]$ is formulated as:
\begin{equation}
\label{eq:nosie_corruption}
    \{\mathbf{x}_t, \mathbf{y}_t\} = \alpha_t \{\mathbf{x}_0, \mathbf{y}_0\}+\sigma_t \bm{\epsilon},
\end{equation}
where \(\alpha_t\) preserves the original signal strength, \(\sigma_t\) controls the intensity of Gaussian noise \(\bm{\epsilon} \sim \mathcal{N}(0, \mathbf{I})\).
However, we observed that the different sensitivities of modalities (SAR images and masks) to noise introduce challenges in the joint generation.
Recent studies \cite{chen_ontheimportance, chen2023generalist} have explored different information loss during the diffusion process for different resolutions and characteristics in a single domain generation, but we, for the first time, tackle imbalanced noise scheduling between two different modalities in the learning of joint generation.
In our case, SAR images \(\mathbf{x}_0\) are represented by continuous pixel values, while segmentation masks \(\mathbf{y}_0\) consist of discrete, binary values.
When noise is added during the generation process, noise corrupts the SAR images and the segmentation masks unevenly, leading to degradation in the quality of the jointly generated outputs.
As shown in \cref{fig:balancing_factor}, the noise degrades the SAR image (the first row) faster than the segmentation mask (the second row) as the timestep increases.

To address this issue, we propose an SNR-based balancing factor to align the noise tolerance between different modalities, improving the fidelity of both generated SAR images and their soft labels.
The noise sensitivity of each modality can be approximated using its signal-to-noise ratio (SNR). 
We derive an SNR-based balancing factor \(b\) based on the ratio between the SNR of SAR images to the SNR of segmentation masks, ensuring the segmentation mask's noise sensitivity is scaled to match that of the SAR images. 
As shown in \cref{fig:balancing_factor}, the segmentation masks scaled with our SNR-based balancing factor (in the second row) show similar noise sensitivity with the corrupted SAR images.
The detailed derivation is provided in the \textit{Supplementary material}.
The balanced noise corruption process for the joint generation of SAR image-label pairs is reformulated as:
\begin{equation}
\label{eq:balanced_noise_corruption}
    \{\mathbf{x}_t, \mathbf{y}_t\} = \alpha_t \{\mathbf{x}_0, b\mathbf{y}_0\}+\sigma_t \bm{\epsilon}.
\end{equation}
It should be noted that \(b\) is calculated for the entire dataset \textit{once a priori}, and is used for training and testing.
The calculated values of \(b\) are 0.5089, 0.4370, and 0.4655 for OSD \cite{related_oilspill_compare_all}, SOS-ALOS \cite{cbdnet}, and SOS-Sentinel \cite{cbdnet}, respectively.

\begin{algorithm}[t]
\footnotesize
\caption{Training Algorithm}
\label{alg: ddpm_training}
\begin{algorithmic}
    \STATE {\bf Input:} \(\mathbf{x}_0 \in \mathbb{R}^{H \times W \times 1}\), \(\Tilde{\mathbf{y}}_0 \in \{0, 1\}^{H \times W \times C} \) 
    \STATE {\bf Output:} loss
    \STATE {\bf Function} \texttt{train\_loss}(\(\mathbf{x}_0\), \( \Tilde{\mathbf{y}}_0 \)):
    \STATE \quad \CommentColorDarkGreen{\texttt{\# Normalize one-hot masks to be -1 or 1}}    
    \STATE \quad \(\mathbf{y}_0 \gets \Tilde{\mathbf{y}}_0 \cdot 2 - 1\)
    \STATE \quad \CommentColorDarkGreen{\texttt{\# Scale with our SNR-based balancing factor} \(b\)}
    
    \STATE \quad \(\mathbf{y}_0' \gets \mathbf{y}_0 \cdot b\)
    \STATE \quad \CommentColorDarkGreen{\texttt{\# Noise corruption process}}
    \STATE \quad $t \sim \text{Uniform}(1, T)$
    \STATE \quad $\bm{\epsilon} \sim \mathcal{N}(0, \mathbf{I})$
    \STATE \quad $(\mathbf{x}_t, \mathbf{y}'_t) \gets \alpha_t \cdot ({\mathbf{x}_0, \mathbf{y}_0'}) + \sigma_t \cdot \bm{\epsilon}$
    \STATE \quad \CommentColorDarkGreen{\texttt{\# Predict $\hat{\mathbf{x}}_\theta, \hat{\mathbf{y}}_\theta$}}
    \STATE \quad $\hat{\mathbf{x}}_\theta, \hat{\mathbf{y}}_\theta \gets \text{DDPM}((\mathbf{x}_t, \mathbf{y}'_t),t)$
    \STATE \quad \CommentColorDarkGreen{\texttt{\# Compute loss}}
    \STATE \quad $\mathcal{L}_{\theta} \gets \text{L2}(\hat{\mathbf{x}}_\theta, \mathbf{x}_0) + \text{Cross\_Entropy}(\hat{\mathbf{y}}_\theta, \Tilde{\mathbf{y}}_0)$
    \STATE \quad \textbf{return} $\mathcal{L}_{_{\theta}}$
\end{algorithmic}
\end{algorithm}

\subsubsection{Training to Synthesize SAR Images and Soft Labels}
\label{sec:train_soft_labels}

\cref{alg: ddpm_training} outlines the DDPM training strategy of our DAKTer to jointly generate SAR images and their soft labels.
As described, the one-hot encoded segmentation mask $\Tilde{\mathbf{y}}_0$ is first normalized and scaled using the proposed SNR-based balancing factor \(b\).
Then, the scaled segmentation mask \(\mathbf{y}'_0\) and SAR image \(\mathbf{x}_0\) are then corrupted with noise at a randomly sampled timestep \(t\).
DDPM jointly predicts the denoised SAR image \(\hat{\mathbf{x}}_\theta\) and logits \(\hat{\mathbf{y}}_\theta\).
For SAR image reconstruction, we use an L2 loss between the denoised SAR image \(\hat{\mathbf{x}}_\theta\) and the ground truth SAR image \(\mathbf{x}_0\).
For soft label generation learning, we use a cross-entropy (CE) loss between the predicted logit \(\hat{\mathbf{y}}_\theta\) and the one-hot segmentation mask $\Tilde{\mathbf{y}}_0$.
Unlike conventional L2 or MSE-based losses, the CE loss encourages logit-based outputs, containing semantically aligned probability values corresponding to their ground truth (GT) labels.
Notably, we \textit{firstly} utilize the CE loss in the generative DA context to generate segmentation masks as soft labels that can be used as supervision for student (downstream) segmentation models.

\begin{algorithm}[t]
\footnotesize
\caption{Inference Algorithm}
\label{alg: inference}

\begin{algorithmic}
    
    
    \STATE \CommentColorDarkGreen{\texttt{\# Initialize}}
    \STATE \(\mathbf{x}_T, \mathbf{y}'_T \sim \mathcal{N}(0, \mathbf{I}) \)
    \FOR{$t = T$ to $1$}
    \STATE \quad  \CommentColorDarkGreen{\texttt{\# Predict $\hat{\mathbf{x}}_\theta, \hat{\mathbf{y}}_\theta$}}
    \STATE \quad $\hat{\mathbf{x}}_\theta, \hat{\mathbf{y}}_\theta \gets \text{DDPM}((\mathbf{x}_t, b\mathbf{y}_t), t)$
    \STATE \quad \CommentColorDarkGreen{\texttt{\# Rescale the predicted logit \(\hat{\mathbf{y}}_\theta\)}}
    \STATE \quad $\hat{\mathbf{y}}_{\theta}' \gets$ \((\text{softmax}(\hat{\mathbf{y}}_\theta) \cdot 2 -1 ) \cdot b\)
    \STATE \quad \CommentColorDarkGreen{\texttt{\# Compute deterministic next step}}
    \STATE \quad $\mathbf{x}_{t-1}, \mathbf{y}'_{t-1} \gets $ DDIM\_step(($\hat{\mathbf{x}}_\theta, \hat{\mathbf{y}}_{\theta}'$), ($\mathbf{x}_t, b\mathbf{y}_t$),  $t$)
    \ENDFOR
    \STATE \CommentColorDarkGreen{\texttt{\# Return SAR image and logit for soft label}}
    \STATE \textbf{return} $\mathbf{x}_0, \hat{\mathbf{y}}_{\theta}$
\end{algorithmic}
\end{algorithm}

\subsection{Synthesizing SAR Images and Soft Labels}
\label{sec:inference}

After training the DDPM (Sec. \ref{sec:learning_joint_distribution}) with our DAKTer strategy, we generate an augmented dataset $\mathcal{D}^a$, consisting of synthetic SAR oil spill images and their corresponding soft labels, following the process outlined in \cref{alg: inference}.
Starting from random Gaussian noise, the DDPM jointly predicts the denoised SAR image \(\hat{\mathbf{x}}_\theta\) and logit \(\hat{\mathbf{y}}_\theta\).
It should be noted that the input noise \(\mathbf{y}_t\) is also scaled by our SNR-based balancing factor \(b\) to align the noise corruption process across modalities.
After applying softmax operation on the predicted logit \(\hat{\mathbf{y}}_\theta\), normalizing and rescaling using \(b\) are applied.
The next input to the DDPM is then computed using the DDIM step operation \cite{song2020denoising}.
Through this inference process, we can generate the augmented dataset $\mathcal{D}^a$ including realistic SAR oil spill images \(\mathbf{x}_0\) and logits \(\hat{\mathbf{y}}_\theta\) used in making soft labels.

\begin{figure*}[t!]
    \centering
    \includegraphics[width=\linewidth]{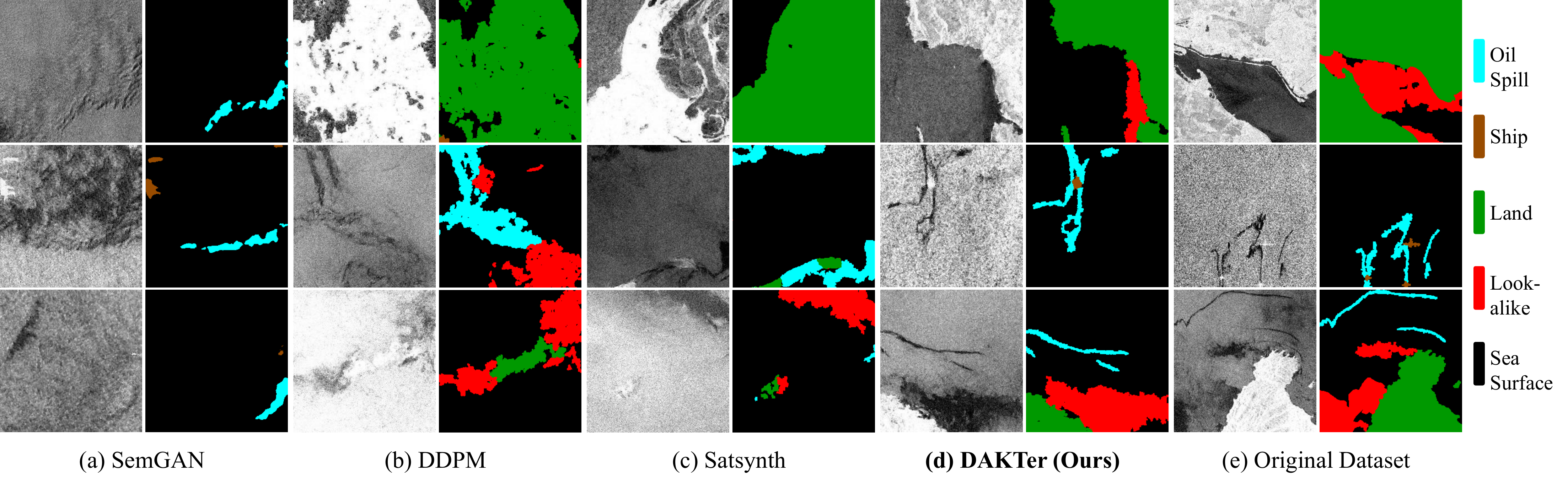}
    \caption{Data generation results: Qualitative comparison of SAR images and corresponding segmentation masks generated by (a) SemGAN \cite{semgan}, (b) DDPM \cite{ddpm}, (c) SatSynth \cite{satsynth}, and (d) Our DAKTer, compared to (e) samples from the original OSD dataset \cite{related_oilspill_compare_all}.}
    \label{fig:OSD_generation_qual}
    \vspace{-3mm}
\end{figure*}

\subsection{Knowledge Transfer with Generated Soft Labels}
\label{sec:kd_using_synthetic_data}

To leverage the augmented dataset $\mathcal{D}^a$, we employ a knowledge distillation loss \cite{hinton2015distillingknowledgeneuralnetwork} to transfer the knowledge of soft labels to student models, such as \cite{cbdnet, exp_deeplabv3+, cheng2021mask2former, experiment_segformer}.
For class $i$, the soft label $\mathbf{q}_{\text{t}}(i)$ is defined as:
\begin{equation}
\label{eq:soft_label}
    \mathbf{q}_{t}(i) = \frac{\exp(\mathbf{\hat{y}}_\theta(i) / T)}{\sum_{j=1}^{C} \exp(\mathbf{\hat{y}}_\theta(j) / T)},
\end{equation}
where $\mathbf{\hat{y}}_\theta$ is generated logit value from our DA model, $T$ is a hyperparameter that softens class probability distribution and $C$ is the number of classes.
Then, the knowledge distillation loss $\mathcal{L}_{\text{kd}}$ is defined as a scaled sum of cross-entropy losses between the soft label $\mathbf{q}_{\text{t}}$ and predicted logit $\mathbf{p}_{\text{s}}$ from a student segmentation network, which is expressed as:
\begin{equation}
    \mathcal{L}_{\text{kd}} = \frac{T^2}{N} \sum_{k=1}^{N} \mathcal{L}_{\text{ce}}(\mathbf{q}_{t}, \mathbf{p}_{s}),
\end{equation}
where \(N\) is the number of pixels.
This knowledge transfer process allows the student segmentation model to learn from the richer and softened outputs (pixel-wise probability maps for predicted segmentation masks) of a diffusion-based DA model, improving the network’s generalizability from our DAKTer strategy.
The student segmentation model is trained with \(\mathcal{L}_{\text{kd}}\) when data is sampled from $\mathcal{D}^a$.
On the other hand, it is trained with $\mathcal{L}_{\text{ce}}$ when data is sampled from $\mathcal{D}$.
Therefore, the loss functions used to train the student segmentation network are expressed as:
\begin{equation} 
     \mathcal{L}_{\text{stu}}^{D^a} = \lambda_{\text{kd}} \mathcal{L}_{\text{kd}} + \lambda_{\text{dice}} \mathcal{L}_{\text{dice}}, 
 \end{equation} 
\begin{equation} 
     \mathcal{L}_{\text{stu}}^{D} = \lambda_{\text{ce}} \mathcal{L}_{\text{ce}} + \lambda_{\text{dice}} \mathcal{L}_{\text{dice}}, 
 \end{equation} 
where $\mathcal{L}_{\text{dice}}$ is a soft dice loss \cite{method_diceloss} that mitigates the class imbalance problems.

\begin{table*}
\scriptsize
\centering
 \setlength{\tabcolsep}{4pt}   
\begin{tabular}{c|l|cc|cccccccc}
    \toprule
     & & \multicolumn{2}{c}{Syn. Quality} &  \multicolumn{2}{c}{DeepLabV3+ \cite{exp_deeplabv3+}} & \multicolumn{2}{c}{CBDNet \cite{cbdnet}} & \multicolumn{2}{c}{Mask2Former \cite{cheng2021mask2former}} & \multicolumn{2}{c}{SegFormer \cite{experiment_segformer}} \\
    \cmidrule(lr){5-12} 
    Dataset & DA Methods & FID ($\downarrow$) & sFID ($\downarrow$) & mIoU (\%)  & F1 (\%) & mIoU (\%) & F1 (\%) & mIoU (\%) & F1 (\%) & mIoU (\%) & F1 (\%)\\
    \midrule
    \multirow{5}{*}{OSD \cite{related_oilspill_compare_all}} & -  & - & - & 65.10  & 76.46 & 66.32 & 77.60 & 65.95 & 77.30 & 67.46 & 78.48 \\
    & SemGAN \cite{semgan} & 36.65 & 9.488 &  66.79 & 78.21 & 67.76 & 79.11 & 66.16 & 77.45 & 67.66 & 79.04 \\  
    & DDPM \cite{ddpm} & 40.85 & 11.15 & 68.61 & 79.77 & 68.58 & 79.68 & 66.84 & 78.02 & 69.77 & 80.61 \\  
    & SatSynth \cite{satsynth} & 39.61 & 10.77 & 68.24 & 79.36 & 68.98 & 80.13 & 66.04 & 77.22 & 69.79 & 80.60 \\  
    & DAKTer (Ours) & \textbf{14.58} & \textbf{6.728} & \textbf{69.35} & \textbf{80.24} & \textbf{70.49} & \textbf{81.24} & \textbf{68.22} & \textbf{79.33} & \textbf{70.74} & \textbf{81.35} \\
    \midrule
    \multirow{5}{*}{SOS-Sentinel \cite{cbdnet}} & -  & - & - & 85.00 & 91.84 & 83.42 & 87.87 & 84.22 & 91.37 & 88.15 & 93.67 \\
    & SemGAN \cite{semgan} & 33.51 & 15.11 & 84.41 & 91.49 & 85.00 &  91.84  & 84.01 &  91.24 & 87.02 & 93.02 \\  
    & DDPM \cite{ddpm} & 29.38 & 11.81 & 84.42 & 91.49 & 85.55 & 92.16 & 84.50 & 91.54 & 87.59 & 93.35  \\  
    & SatSynth \cite{satsynth} & 23.58 & \textbf{11.01} & 84.40 & 91.48 & 85.47 & 92.11 & 84.26 & 91.40 & 87.98 & 93.57  \\  
    & DAKTer (Ours) & \textbf{22.43} & 11.03 & \textbf{85.39} & \textbf{92.07} & \textbf{86.04} & \textbf{92.45} & \textbf{85.13} & \textbf{91.91} & \textbf{88.74} & \textbf{94.00} \\
    \midrule
    \multirow{5}{*}{SOS-ALOS \cite{cbdnet}} & -  & - & - & 84.15 & 91.08 & 83.31 & 84.84 & 83.80 & 90.86 & 86.62 & 92.61 \\
    & SemGAN \cite{semgan} & 61.72 & 38.83 & 83.95 & 90.95 & 84.68 & 91.41 & 83.60 & 90.73 & 86.88 & 92.77  \\  
    & DDPM \cite{ddpm}  & 43.94 & 15.04 & 84.20 & 91.10 & 84.77 & 91.46  & 83.88 & 90.90 & 86.78 & 92.70  \\  
    & SatSynth \cite{satsynth} & 32.73 & 13.85 & 84.30 & 91.17  & 84.61 & 91.37  & 84.01 & 90.99 & 86.74 & 92.68 \\  
    & DAKTer (Ours) & \textbf{30.23} &  \textbf{13.05} &  \textbf{84.64} & \textbf{91.38}  & \textbf{85.21} & \textbf{91.74} & \textbf{84.16} & \textbf{91.08} & \textbf{87.52} & \textbf{93.16} \\
    \bottomrule
\end{tabular}
\caption{Comparison of data augmentation methods on OSD \cite{related_oilspill_compare_all}, SOS-Sentinel, and SOS-ALOS \cite{cbdnet} dataset. We evaluate the quality of the synthesized SAR images at a resolution of \(256\times256\) using FID \cite{fid} and sFID \cite{sfid} metrics. Also, we compare segmentation performance by training recent segmentation models \cite{exp_deeplabv3+, cbdnet, cheng2021mask2former, experiment_segformer} on the original dataset and 100\% augmented datasets synthesized by SemGAN \cite{semgan}, DDPM \cite{ddpm}, SatSynth \cite{satsynth}, and our DAKTer.}
\label{tab:dakd_osd}
\vspace{-3mm}
\end{table*}
\section{Experiments}

\subsection{Datasets}

\textbf{Oil Spill Detection (OSD) dataset} \cite{related_oilspill_compare_all} was constructed using SAR images from the Sentinel-1 mission.
Each image contains annotated instances across five semantic classes: `oil-spill', `look-alike', `land', `ship', and `sea surface'. 
It consists of 1,002 images for training and 110 images for testing at a resolution of \(650\times1,250\).
We crop the SAR images and their masks to a resolution of \(256\times256\), resulting in \(8,760\) pairs for training and 948 pairs for testing.

\noindent\textbf{Deep-SAR Oil Spill (SOS) dataset} \cite{cbdnet} has two splits: SAR images captured by the PALSAR sensor on the ALOS satellite, and SAR images captured by Sentinel-1A satellite. 
The dataset contains 8,070 labeled SAR images of $256\times256$ resolution, cropped from 21 raw SAR images.
For the ALOS dataset (SOS-ALOS), 3,101 images are allocated for training, and 776 images are used for testing. 
Similarly, for the Sentinel-1A dataset (SOS-Sentinel), 3,354 images are used for training, and 839 images are used for testing.

\subsection{Comparison with other DA methods}
\subsubsection{SAR Image Generation Quality}
First, we compare the data generation quality of our DAKTer strategy (ours) against other generative DA methods, including SemGAN \cite{semgan}, DDPM \cite{ddpm}, and SatSynth \cite{satsynth}.
For SatSynth \cite{satsynth}, we follow to encode segmentation masks using binary encoding.
For training label generation, both comparisons of DDPM \cite{ddpm} and SatSynth \cite{satsynth} employ L2 loss, whereas our DAKTer strategy utilizes the CE loss, enabling logit-based soft label generation.
For a fair comparison, we use the same DDPM architecture size to reproduce diffusion-based DA methods \cite{ddpm, satsynth} as ours and train all models with the same number of iterations.

\textbf{Quantitative comparision}
As shown in `Syn. Quality' column in Table \ref{tab:dakd_osd}, our DAKTer consistently outperforms other DA methods. 
Specifically, on the SAR images of OSD dataset \cite{related_oilspill_compare_all}, our DAKTer marks \(14.58\) and \(6.728\), in terms of FID \cite{fid} and sFID \cite{sfid}, respectively.
Our SNR-based balancing factor helps DDPM synthesize more realistic SAR oil spill images compared to other diffusion-based DA methods \cite{ddpm, satsynth}.
We further ablate the effect of the SNR-based balancing factor in \cref{sec:abl_balancing_factor}.

\textbf{Qualitative comparison}. 
\cref{fig:OSD_generation_qual} shows the generated SAR oil spill images and their corresponding segmentation masks.
It should be noted that our segmentation masks are thresholded from the generated soft labels.
SemGAN \cite{semgan} fails to generate diverse SAR images and segmentation masks due to the widely known mode collapse problems of GAN \cite{arjovsky2017wasserstein}.
DDPM \cite{ddpm} and Satsynth \cite{satsynth} often generate poorly aligned pairs of SAR image and mask and non-realistic regions such as `sea surface' regions inside `land' regions (the first row) and `land' regions inside `oil-spill' regions (second row).
Also, in the last row, the SAR images from (b) and (c) are extremely white and highly saturated.
DAKTer-generated images exhibit balanced brightness levels and well-aligned segmentation masks, accurately capturing the boundaries of `oil spill' and `look-alike' regions.
Additionally, our DAKTer accurately generates `oil-spill' and `look-alike' regions as dark while preserving each characteristic from the original dataset, including the blurred and broad contours for `look-alike' regions and the long and sharp contrast for `oil spill' regions.

\subsubsection{Segmentation Performance}
To compare segmentation performance improvement from other DA methods against our DAKTer, we train four recent semantic segmentation models \cite{exp_deeplabv3+, cbdnet, cheng2021mask2former, experiment_segformer} using the augmented datasets from \cite{semgan, ddpm, satsynth} and our DAKTer with the original datasets.
Notably, the augmented data from \cite{semgan, ddpm, satsynth} contain hard labels, whereas our DAKTer strategy utilizes soft labels for training segmentation models. 
Also, the amount of augmented data generated for each method is equal to the original dataset size.

As shown in \cref{tab:dakd_osd}, for the OSD dataset \cite{related_oilspill_compare_all}, all segmentation models exhibit only marginal improvements when using the GAN-based DA method \cite{semgan}. 
Although SatSynth \cite{satsynth} shows better visual quality in terms of FID and sFID than DDPM \cite{ddpm}, the superior segmentation performance from DDPM comes from the misalignment between the SAR images and the mask images as shown in \cref{fig:OSD_generation_qual}
However, our proposed DAKTer achieves the most significant enhancement, outperforming all other data augmentation methods, thanks to its ability to generate soft labels that contain richer semantic information for training. 
Furthermore, for SOS-Sentinel \& ALOS \cite{cbdnet}, other DA methods struggle to improve the performance of segmentation models \cite{exp_deeplabv3+, cbdnet, cheng2021mask2former, experiment_segformer}.
On the other hand, our DAKTer consistently boosts the performance of segmentation models \cite{exp_deeplabv3+, cbdnet, cheng2021mask2former, experiment_segformer}, achieving the highest scores in all metrics.
This highlights that our DAKTer strategy effectively handles data scarcity problems in SAR oil spill segmentation.

\begin{table}
    \scriptsize
    \centering
    \begin{tabular}{cccccccc}
    \toprule
    Exp. & DA Method & Teacher & Student & mIoU (\%) & F1 (\%)  \\
    \midrule
    (a) & - & - & SF-B3 & 67.46 & 78.48 \\
    (b) & - & SF-B5 &  & 70.06 & 81.05 \\
    (c) & - & SF-B5 & SF-B3 & 69.93 & 80.88 \\
    \midrule
    (d) & DDPM \cite{ddpm} &  &  SF-B3 & 69.77 & 80.61 \\
    (e) & DDPM \cite{ddpm} & SF-B5 &  & 70.07 & 80.86 \\
    (f) & DDPM \cite{ddpm} & SF-B5 & SF-B3  & 69.36 & 80.48 \\
    \midrule
    (g) & \multicolumn{2}{c}{Ours}  & SF-B3 & \textbf{70.74} & \textbf{81.35} \\
    \bottomrule
\end{tabular}

    \caption{Comparison with other knowledge distillation methods using soft labels on the OSD dataset \cite{related_oilspill_compare_all}. SF-B3 and SF-B5 refer to as SegFormer-B3 and SegFormer-B5 \cite{experiment_segformer}, respectively.}
    \label{tab:kd_osd}
    \vspace{-3mm}
\end{table}

\subsection{Comparison with KD methods}

To further demonstrate the knowledge transfer capability of our DAKTer, we compare it with other knowledge distillation (KD) methods in Table \ref{tab:kd_osd} where:

\noindent\textbf{Exp-(a-c):} 
We apply a KD method \cite{hinton2015distillingknowledgeneuralnetwork} for training a student model (SF-B3) using precomputed soft labels from a larger teacher model (SF-B5).
As SF-B5 has twice as many network parameters (82.0M) than SF-B3 (45.2M), teacher (Exp-a) outperforms student model (Exp-b) by $2.6\%$ points on mIoU.
While the student model shows notable performance improvement, the effect remains limited, likely due to the insufficient training data available for both the teacher and student models (Exp-c).

\noindent\textbf{Exp-(d-f):} Inspired by the approach of Chen \etal \cite{Chen_2019_ICCV}, we experiment simple combination of generative DA and KD.
We first train DDPM \cite{ddpm} to generate an augmented dataset of SAR images and one-hot labels.
Then, we train a teacher model using the augmented dataset with the original dataset (Exp-e).
Using the teacher model, we compute soft labels for the augmented dataset, which is then used with the original dataset to train the student model (Exp-f).
While this approach (Exp-f) improves performance, it remains inferior to directly augmenting the dataset (Exp-d).
This performance gap arises from the two-stage process (DA followed by teacher training), where errors originating from the augmented data are amplified during soft label prediction, reducing their reliability.

\noindent\textbf{Exp-(g):} We train the student model using our DAKTer strategy.
Unlike previous experiments, our DAKTer does not require an additional teacher model trained for segmentation tasks. Nevertheless, the student model achieves the highest mIoU and F1 scores, demonstrating that the synthesized soft labels from DDPM trained with our DAKTer effectively enhance segmentation performance.

\subsection{Effect of SNR-based Balancing Factor}
\label{sec:abl_balancing_factor}

\begin{table}
\scriptsize
\centering
\setlength{\tabcolsep}{7pt} 
\begin{tabular}{c | ccc c}
\toprule
& \multicolumn{2}{c}{Syn. Quality} & \multicolumn{2}{c}{SegFormer \cite{experiment_segformer}} \\
 \(b\)    & FID ($\downarrow$) & sFID ($\downarrow$) & mIoU (\%) & F1 (\%)\\ 
\midrule
0.1  & 25.71 & 10.01 & 69.05 & 79.83 \\
0.5089 (Ours)& \textbf{14.58} & \textbf{6.728} & \textbf{70.74} & \textbf{81.35} \\
1 & 20.38  & 7.704 & 69.72 & 80.65 \\
2 &  28.47 & 9.571 & 70.00 & 80.75 \\
\bottomrule
\end{tabular}

\caption{Ablation analysis on the balancing factor $b$. We assess the quality of generated SAR images and the segmentation performance of SegFormer \cite{experiment_segformer} trained with our DAKTer strategy and evaluated on the OSD dataset \cite{related_oilspill_compare_all}.}
\label{tab:abl_b_suppl}
\vspace{-3mm}
\end{table}

We assess the effectiveness of our proposed SNR-based balancing factor \(b\) in the joint generation.
In \cref{tab:abl_b_suppl}, we first compare the generation quality of SAR images from DDPM models trained and tested with different \(b\) values.
Furthermore, we evaluate the segmentation performance of SegFormer \cite{experiment_segformer} trained on augmentation datasets including SAR images and soft labels generated from each case.

For the OSD dataset \cite{related_oilspill_compare_all}, the calculated optimal value of \(b\) based on the SNR is 0.5089.
For comparison, we also trained and tested DDPM with \(b=0.1, 1\), and 2.
When \(b\) is too small (e.g., 0.1), the signal of the segmentation mask becomes too weak compared to SAR images, while larger values (e.g., 1 or 2) over-amplify it.
Results show that DDPM trained with \(b=0.5089\) synthesizes more realistic SAR images compared to other variants.
Also, the segmentation model, trained with the augmented dataset from the DDPM with \(b=0.5089\), consistently outperforms other settings.
These results highlight that our SNR-based balancing factor effectively stabilizes joint generation, ensuring a well-balanced synthesis of SAR images and their soft labels.

\begin{figure}[t!]
    \centering
    \includegraphics[width=0.8\linewidth]{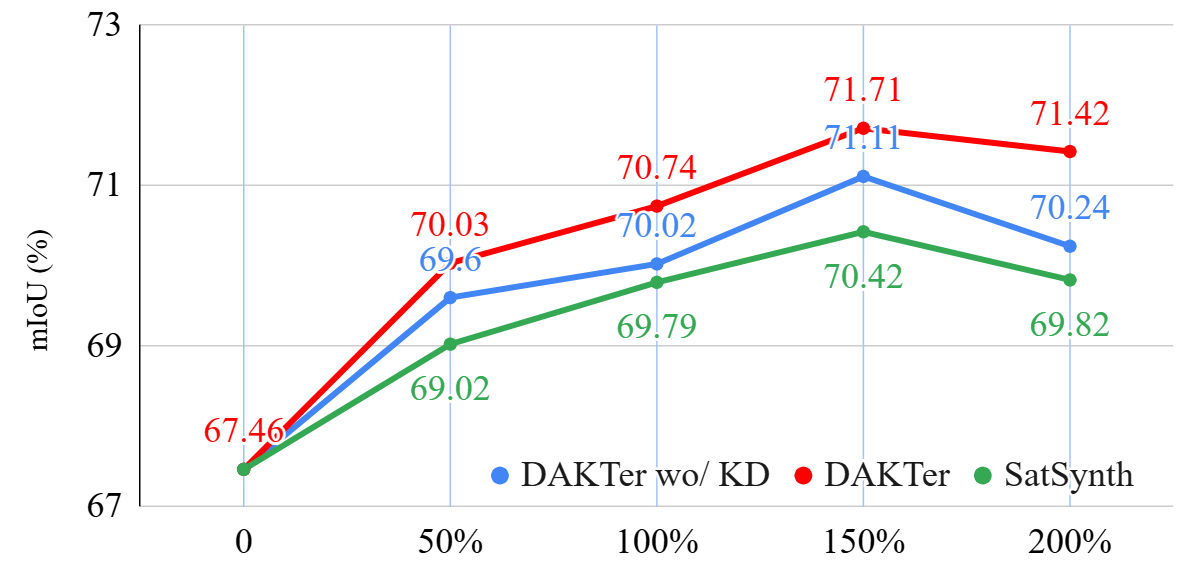}
    \caption{Comparsion of segmentation performance on \cite{related_oilspill_compare_all} across different augmentation scales (0, $50\%$, $100\%$, $150\%$, and $200\%$ of the original training set \cite{related_oilspill_compare_all}) for training SegFormer \cite{experiment_segformer} with SatSynth \cite{satsynth}, our DAKTer without KD (soft-label supervision), and our full DAKTer strategy.}
    \label{fig:scale_osd}
\end{figure}

\subsection{Effect from Scale of Synthetic Data}
We evaluate the impact of different scales of augmented datasets on segmentation performance. 
Specifically, we apply our DAKTer strategy at $50\%$, $100\%$, $150\%$, and $200\%$ augmentation scales and measure segmentation performance using SegFormer \cite{experiment_segformer}.
For comparison, we also generate datasets at the same scales using SatSynth \cite{satsynth}.
To further assess the contribution of generated soft labels, we measure the performance using DAKTer without soft labels to supervise student models (DAKTer wo/ KD).
As shown in \cref{fig:scale_osd}, segmentation performances of our DAKTer strategy consistently outperform in all scales compared to SatSynth \cite{satsynth} and DAKTer without soft label supervision.
These results demonstrate that our DAKTer strategy not only improves segmentation performance but also ensures greater robustness across varying augmentation scales by leveraging the rich semantic information in soft labels.
\section{Conclusion}
\label{conclusion}

In this work, we proposed a diffusion-based data augmentation and knowledge transfer strategy (DAKTer) to address the scarcity of labeled SAR oil spill data. 
Our approach leverages DDPM to jointly generate realistic SAR oil spill images and soft labels, which contain per-pixel 
class probability, enabling more effective supervision for segmentation models. 
We introduced an SNR-based balancing mechanism that aligns noise corruption across SAR images and segmentation masks, stabilizing joint generation.
Extensive experiments demonstrated that our DAKTer strategy significantly enhances the performance of student segmentation models, outperforming generative DA methods. 
Our results highlight that diffusion-generated soft labels provide richer semantic information than one-hot labels, facilitating better knowledge transfer to student models.
Our DAKTer strategy has enhanced AI-driven environmental monitoring and shown potential for integrating into real-world marine surveillance systems to aid disaster response teams, demonstrating its feasibility in detecting oil spills.

\newpage
{
    \small
    \bibliographystyle{ieeenat_fullname}
    \bibliography{main}

\begin{thebibliography}{75}
\providecommand{\natexlab}[1]{#1}
\providecommand{\url}[1]{\texttt{#1}}
\expandafter\ifx\csname urlstyle\endcsname\relax
  \providecommand{\doi}[1]{doi: #1}\else
  \providecommand{\doi}{doi: \begingroup \urlstyle{rm}\Url}\fi

\bibitem[Albanie et~al.(2018)Albanie, Nagrani, Vedaldi, and Zisserman]{multi-modal_kd1}
Samuel Albanie, Arsha Nagrani, Andrea Vedaldi, and Andrew Zisserman.
\newblock Emotion recognition in speech using cross-modal transfer in the wild.
\newblock In \emph{Proceedings of the 26th ACM International Conference on Multimedia}, page 292–301, New York, NY, USA, 2018. Association for Computing Machinery.

\bibitem[Arjovsky et~al.(2017)Arjovsky, Chintala, and Bottou]{arjovsky2017wasserstein}
Martin Arjovsky, Soumith Chintala, and L{\'e}on Bottou.
\newblock Wasserstein generative adversarial networks.
\newblock In \emph{International conference on machine learning}, pages 214--223. PMLR, 2017.

\bibitem[Azizi et~al.(2023)Azizi, Kornblith, Saharia, Norouzi, and Fleet]{intro21_da_wo_syndataimprove_clss}
Shekoofeh Azizi, Simon Kornblith, Chitwan Saharia, Mohammad Norouzi, and David~J Fleet.
\newblock Synthetic data from diffusion models improves imagenet classification.
\newblock \emph{arXiv preprint arXiv:2304.08466}, 2023.

\bibitem[Baranchuk et~al.(2021)Baranchuk, Rubachev, Voynov, Khrulkov, and Babenko]{label-efficient_seg_diff_mlp}
Dmitry Baranchuk, Ivan Rubachev, Andrey Voynov, Valentin Khrulkov, and Artem Babenko.
\newblock Label-efficient semantic segmentation with diffusion models.
\newblock \emph{arXiv preprint arXiv:2112.03126}, 2021.

\bibitem[Bianchi et~al.(2020)Bianchi, Espeseth, and Borch]{related_oilspill_unet}
F.~M. Bianchi, M.~M. Espeseth, and N. Borch.
\newblock Large-scale detection and categorization of oil spills from sar images with deep learning.
\newblock \emph{Remote Sensing}, 12\penalty0 (14):\penalty0 2260, 2020.

\bibitem[Chen et~al.(2019)Chen, Wang, Xu, Yang, Liu, Shi, Xu, Xu, and Tian]{Chen_2019_ICCV}
Hanting Chen, Yunhe Wang, Chang Xu, Zhaohui Yang, Chuanjian Liu, Boxin Shi, Chunjing Xu, Chao Xu, and Qi Tian.
\newblock Data-free learning of student networks.
\newblock In \emph{Proceedings of the IEEE/CVF International Conference on Computer Vision (ICCV)}, 2019.

\bibitem[Chen et~al.(2023{\natexlab{a}})Chen, Xie, Chen, Wang, Hong, Li, and Yeung]{GEODIFFUSION}
Kai Chen, Enze Xie, Zhe Chen, Yibo Wang, Lanqing Hong, Zhenguo Li, and Dit-Yan Yeung.
\newblock Geodiffusion: Text-prompted geometric control for object detection data generation.
\newblock \emph{arXiv preprint arXiv:2306.04607}, 2023{\natexlab{a}}.

\bibitem[Chen et~al.(2018)Chen, Zhu, Papandreou, Schroff, and Adam]{exp_deeplabv3+}
Liang-Chieh Chen, Yukun Zhu, George Papandreou, Florian Schroff, and Hartwig Adam.
\newblock Encoder-decoder with atrous separable convolution for semantic image segmentation.
\newblock In \emph{Proceedings of the European conference on computer vision (ECCV)}, pages 801--818, 2018.

\bibitem[Chen et~al.(2021)Chen, Liu, Zhao, and Jia]{reviewkd}
Pengguang Chen, Shu Liu, Hengshuang Zhao, and Jiaya Jia.
\newblock Distilling knowledge via knowledge review.
\newblock In \emph{IEEE Conference on Computer Vision and Pattern Recognition (CVPR)}, 2021.

\bibitem[Chen(2023)]{chen_ontheimportance}
Ting Chen.
\newblock On the importance of noise scheduling for diffusion models.
\newblock \emph{arXiv preprint arXiv:2301.10972}, 2023.

\bibitem[Chen et~al.(2023{\natexlab{b}})Chen, Li, Saxena, Hinton, and Fleet]{chen2023generalist}
Ting Chen, Lala Li, Saurabh Saxena, Geoffrey Hinton, and David~J Fleet.
\newblock A generalist framework for panoptic segmentation of images and videos.
\newblock In \emph{Proceedings of the IEEE/CVF international conference on computer vision}, pages 909--919, 2023{\natexlab{b}}.

\bibitem[Chen et~al.(2020)Chen, Li, and Wang]{related_oilspill_deeplab}
Y. Chen, Y. Li, and J. Wang.
\newblock An end-to-end oil-spill monitoring method for multisensory satellite images based on deep semantic segmentation.
\newblock \emph{Sensors}, 20\penalty0 (3):\penalty0 725, 2020.

\bibitem[Cheng et~al.(2022)Cheng, Misra, Schwing, Kirillov, and Girdhar]{cheng2021mask2former}
Bowen Cheng, Ishan Misra, Alexander~G Schwing, Alexander Kirillov, and Rohit Girdhar.
\newblock Masked-attention mask transformer for universal image segmentation.
\newblock In \emph{Proceedings of the IEEE/CVF conference on computer vision and pattern recognition}, pages 1290--1299, 2022.

\bibitem[Choi and Jeong(2019)]{intro_speckle_noise_jh}
Hyunho Choi and Jechang Jeong.
\newblock Speckle noise reduction technique for sar images using statistical characteristics of speckle noise and discrete wavelet transform.
\newblock \emph{Remote Sensing}, 11\penalty0 (10), 2019.

\bibitem[Contributors(2022)]{mmyolo2022}
MMYOLO Contributors.
\newblock {MMYOLO: OpenMMLab YOLO} series toolbox and benchmark.
\newblock \url{https://github.com/open-mmlab/mmyolo}, 2022.

\bibitem[Do et~al.(2019)Do, Do, Tran, Tjiputra, and Tran]{multi-modal_kd3}
Tuong Do, Thanh-Toan Do, Huy Tran, Erman Tjiputra, and Quang~D. Tran.
\newblock Compact trilinear interaction for visual question answering.
\newblock In \emph{Proceedings of the IEEE/CVF International Conference on Computer Vision (ICCV)}, 2019.

\bibitem[Espeseth et~al.(2020)Espeseth, Jones, Holt, Brekke, and Skrunes]{intro2_SARimage_description}
Martine Espeseth, Cathleen Jones, Benjamin Holt, Camilla Brekke, and Stine Skrunes.
\newblock Oil-spill-response-oriented information products derived from a rapid-repeat time series of sar images.
\newblock \emph{IEEE Journal of Selected Topics in Applied Earth Observations and Remote Sensing}, PP:\penalty0 1--1, 2020.

\bibitem[Fang et~al.(2022)Fang, Mo, Wang, Song, Bei, Zhang, and Song]{kd:up_to_100x}
Gongfan Fang, Kanya Mo, Xinchao Wang, Jie Song, Shitao Bei, Haofei Zhang, and Mingli Song.
\newblock Up to 100$\times$ faster data-free knowledge distillation.
\newblock \emph{arXiv preprint arXiv:2112.06253}, 2022.

\bibitem[Fang et~al.(2024)Fang, Han, Zhang, Zhou, Hu, and Ye]{intro16_DA_label_object_ctrldiff}
Haoyang Fang, Boran Han, Shuai Zhang, Su Zhou, Cuixiong Hu, and Wen-Ming Ye.
\newblock Data augmentation for object detection via controllable diffusion models.
\newblock In \emph{Proceedings of the IEEE/CVF Winter Conference on Applications of Computer Vision (WACV)}, pages 1257--1266, 2024.

\bibitem[Fu et~al.(2024)Fu, Chen, Qiao, and Yu]{da_label_clss_dreamda}
Yunxiang Fu, Chaoqi Chen, Yu Qiao, and Yizhou Yu.
\newblock Dreamda: Generative data augmentation with diffusion models.
\newblock \emph{arXiv preprint arXiv:2403.12803}, 2024.

\bibitem[Gemme and Dellepiane(2018)]{intro5_oilspill}
L. Gemme and S.~G. Dellepiane.
\newblock An automatic data-driven method for sar image segmentation in sea surface analysis.
\newblock \emph{IEEE Transactions on Geoscience and Remote Sensing}, 56\penalty0 (5):\penalty0 2633--2646, 2018.

\bibitem[Goodfellow et~al.(2020)Goodfellow, Pouget-Abadie, Mirza, Xu, Warde-Farley, Ozair, Courville, and Bengio]{goodfellow2020gan}
Ian Goodfellow, Jean Pouget-Abadie, Mehdi Mirza, Bing Xu, David Warde-Farley, Sherjil Ozair, Aaron Courville, and Yoshua Bengio.
\newblock Generative adversarial networks.
\newblock \emph{Communications of the ACM}, 63\penalty0 (11):\penalty0 139--144, 2020.

\bibitem[Guo et~al.(2023)Guo, Yan, Li, and Lin]{CAT}
Ziyao Guo, Haonan Yan, Hui Li, and Xiaodong Lin.
\newblock Class attention transfer based knowledge distillation.
\newblock In \emph{Proceedings of the IEEE/CVF Conference on Computer Vision and Pattern Recognition}, pages 11868--11877, 2023.

\bibitem[Heusel et~al.(2017)Heusel, Ramsauer, Unterthiner, Nessler, and Hochreiter]{fid}
Martin Heusel, Hubert Ramsauer, Thomas Unterthiner, Bernhard Nessler, and Sepp Hochreiter.
\newblock Gans trained by a two time-scale update rule converge to a local nash equilibrium.
\newblock \emph{Advances in neural information processing systems}, 30, 2017.

\bibitem[Hinton et~al.(2015)Hinton, Vinyals, and Dean]{hinton2015distillingknowledgeneuralnetwork}
Geoffrey Hinton, Oriol Vinyals, and Jeff Dean.
\newblock Distilling the knowledge in a neural network.
\newblock \emph{arXiv preprint arXiv:1503.02531}, 2015.

\bibitem[Ho et~al.(2020)Ho, Jain, and Abbeel]{ddpm}
Jonathan Ho, Ajay Jain, and Pieter Abbeel.
\newblock Denoising diffusion probabilistic models.
\newblock \emph{Advances in neural information processing systems}, 33:\penalty0 6840--6851, 2020.

\bibitem[Kirillov et~al.(2023)Kirillov, Mintun, Ravi, Mao, Rolland, Gustafson, Xiao, Whitehead, Berg, Lo, et~al.]{related_sam}
Alexander Kirillov, Eric Mintun, Nikhila Ravi, Hanzi Mao, Chloe Rolland, Laura Gustafson, Tete Xiao, Spencer Whitehead, Alexander~C Berg, Wan-Yen Lo, et~al.
\newblock Segment anything.
\newblock In \emph{Proceedings of the IEEE/CVF International Conference on Computer Vision}, pages 4015--4026, 2023.

\bibitem[Konik and Bradtke(2016)]{intro1_SARimage_description}
M. Konik and K. Bradtke.
\newblock Object-oriented approach to oil spill detection using envisat asar images.
\newblock \emph{ISPRS Journal of Photogrammetry and Remote Sensing}, 118:\penalty0 37--52, 2016.

\bibitem[Konz et~al.(2024)Konz, Chen, Dong, and Mazurowski]{intro20_da_label_anatomic_ctr_med_seg}
Nicholas Konz, Yuwen Chen, Haoyu Dong, and Maciej~A Mazurowski.
\newblock Anatomically-controllable medical image generation with segmentation-guided diffusion models.
\newblock In \emph{International Conference on Medical Image Computing and Computer-Assisted Intervention}, pages 88--98. Springer, 2024.

\bibitem[Krestenitis et~al.(2019)Krestenitis, Orfanidis, Ioannidis, Avgerinakis, Vrochidis, and Kompatsiaris]{related_oilspill_compare_all}
M. Krestenitis, G. Orfanidis, K. Ioannidis, K. Avgerinakis, S. Vrochidis, and I. Kompatsiaris.
\newblock Oil spill identification from satellite images using deep neural networks.
\newblock \emph{Remote Sensing}, 11\penalty0 (15):\penalty0 1762, 2019.

\bibitem[Lee et~al.(2016)Lee, Shamsoddini, Li, Trinder, and Li]{intro9_SARimage_description}
I.~K. Lee, A. Shamsoddini, X. Li, J.~C. Trinder, and Z. Li.
\newblock Extracting hurricane eye morphology from spaceborne sar images using morphological analysis.
\newblock \emph{ISPRS Journal of Photogrammetry and Remote Sensing}, 117:\penalty0 115--125, 2016.

\bibitem[Li et~al.(2021{\natexlab{a}})Li, Yang, Kreis, Torralba, and Fidler]{semgan}
Daiqing Li, Junlin Yang, Karsten Kreis, Antonio Torralba, and Sanja Fidler.
\newblock Semantic segmentation with generative models: Semi-supervised learning and strong out-of-domain generalization.
\newblock In \emph{Proceedings of the IEEE/CVF Conference on Computer Vision and Pattern Recognition}, pages 8300--8311, 2021{\natexlab{a}}.

\bibitem[Li et~al.(2021{\natexlab{b}})Li, Ye, Song, Huang, and Pan]{li2021online}
Zheng Li, Jingwen Ye, Mingli Song, Ying Huang, and Zhigeng Pan.
\newblock Online knowledge distillation for efficient pose estimation.
\newblock In \emph{Proceedings of the IEEE/CVF international conference on computer vision}, pages 11740--11750, 2021{\natexlab{b}}.

\bibitem[Li et~al.(2023)Li, Li, Zhao, Song, Li, and Yang]{is_synth_data_ready_for_KD}
Zheng Li, Yuxuan Li, Penghai Zhao, Renjie Song, Xiang Li, and Jian Yang.
\newblock Is synthetic data from diffusion models ready for knowledge distillation?
\newblock \emph{arXiv preprint arXiv:2305.12954}, 2023.

\bibitem[Liu et~al.(2019)Liu, Chen, Liu, Qin, Luo, and Wang]{liu2019structured}
Yifan Liu, Ke Chen, Chris Liu, Zengchang Qin, Zhenbo Luo, and Jingdong Wang.
\newblock Structured knowledge distillation for semantic segmentation.
\newblock In \emph{Proceedings of the IEEE/CVF conference on computer vision and pattern recognition}, pages 2604--2613, 2019.

\bibitem[Loshchilov(2017)]{adamW}
I Loshchilov.
\newblock Decoupled weight decay regularization.
\newblock \emph{arXiv preprint arXiv:1711.05101}, 2017.

\bibitem[Nieto-Hidalgo et~al.(2018)Nieto-Hidalgo, Gallego, Gil, and Pertusa]{intro6_oilspill}
M. Nieto-Hidalgo, A.~J. Gallego, P. Gil, and A. Pertusa.
\newblock Two-stage convolutional neural network for ship and spill detection using slar images.
\newblock \emph{IEEE Transactions on Geoscience and Remote Sensing}, 56\penalty0 (9):\penalty0 5217--5230, 2018.

\bibitem[Park et~al.(2019)Park, Kim, Lu, and Cho]{park2019relational}
Wonpyo Park, Dongju Kim, Yan Lu, and Minsu Cho.
\newblock Relational knowledge distillation.
\newblock In \emph{Proceedings of the IEEE/CVF conference on computer vision and pattern recognition}, pages 3967--3976, 2019.

\bibitem[Rombach et~al.(2022)Rombach, Blattmann, Lorenz, Esser, and Ommer]{rombach2022high}
Robin Rombach, Andreas Blattmann, Dominik Lorenz, Patrick Esser, and Bj{\"o}rn Ommer.
\newblock High-resolution image synthesis with latent diffusion models.
\newblock In \emph{Proceedings of the IEEE/CVF conference on computer vision and pattern recognition}, pages 10684--10695, 2022.

\bibitem[Romero et~al.(2014)Romero, Ballas, Kahou, Chassang, Gatta, and Bengio]{fitnet}
Adriana Romero, Nicolas Ballas, Samira~Ebrahimi Kahou, Antoine Chassang, Carlo Gatta, and Y. Bengio.
\newblock Fitnets: Hints for thin deep nets.
\newblock \emph{arXiv}, 2014.

\bibitem[Salberg et~al.(2014)Salberg, Rudjord, and Solberg]{intro3_darkspot}
A.-B. Salberg, O. Rudjord, and A.~H.~S. Solberg.
\newblock Oil spill detection in hybrid-polarimetric sar images.
\newblock \emph{IEEE Transactions on Geoscience and Remote Sensing}, 52\penalty0 (10):\penalty0 6521--6533, 2014.

\bibitem[Schnell et~al.(2024)Schnell, Wang, Qi, Hu, and Tang]{ScribbleGen}
Jacob Schnell, Jieke Wang, Lu Qi, Vincent~Tao Hu, and Meng Tang.
\newblock Generative data augmentation improves scribble-supervised semantic segmentation.
\newblock In \emph{CVPR 2024 Workshop SyntaGen: Harnessing Generative Models for Synthetic Visual Datasets}, 2024.

\bibitem[Shanmukh and Priya(2024)]{related_oilspill_dualattention}
M~Phani Shanmukh and S~Baghavathi Priya.
\newblock Enhancing oil spill detection using synthetic aperture radar with dual attention u-net model.
\newblock In \emph{2024 15th International Conference on Computing Communication and Networking Technologies (ICCCNT)}, pages 1--6, 2024.

\bibitem[Shu et~al.(2010)Shu, Li, Yousif, and Gomes]{intro7_oilspill_lookalike}
Y. Shu, J. Li, H. Yousif, and G. Gomes.
\newblock Dark-spot detection from sar intensity imagery with spatial density thresholding for oil-spill monitoring.
\newblock \emph{Remote Sensing of Environment}, 114\penalty0 (9):\penalty0 2026--2035, 2010.

\bibitem[Singh and Shree(2016)]{intro15_specklenoise}
Prabhishek Singh and Raj Shree.
\newblock Analysis and effects of speckle noise in sar images.
\newblock In \emph{2016 2nd International Conference on Advances in Computing, Communication, \& Automation (ICACCA) (Fall)}, pages 1--5, 2016.

\bibitem[Song et~al.(2020)Song, Meng, and Ermon]{song2020denoising}
Jiaming Song, Chenlin Meng, and Stefano Ermon.
\newblock Denoising diffusion implicit models.
\newblock \emph{arXiv preprint arXiv:2010.02502}, 2020.

\bibitem[Sudre et~al.(2017)Sudre, Li, Vercauteren, Ourselin, and Jorge~Cardoso]{method_diceloss}
Carole~H. Sudre, Wenqi Li, Tom Vercauteren, Sebastien Ourselin, and M. Jorge~Cardoso.
\newblock \emph{Generalised Dice Overlap as a Deep Learning Loss Function for Highly Unbalanced Segmentations}, page 240–248.
\newblock Springer International Publishing, 2017.

\bibitem[Szegedy et~al.(2016)Szegedy, Vanhoucke, Ioffe, Shlens, and Wojna]{sfid}
Christian Szegedy, Vincent Vanhoucke, Sergey Ioffe, Jon Shlens, and Zbigniew Wojna.
\newblock Rethinking the inception architecture for computer vision.
\newblock In \emph{Proceedings of the IEEE Conference on Computer Vision and Pattern Recognition (CVPR)}, 2016.

\bibitem[{Temitope Yekeen} et~al.(2020){Temitope Yekeen}, Balogun, and {Wan Yusof}]{related_oilspill_maskrcnn}
Shamsudeen {Temitope Yekeen}, Abdul‐Lateef Balogun, and Khamaruzaman~B. {Wan Yusof}.
\newblock A novel deep learning instance segmentation model for automated marine oil spill detection.
\newblock \emph{ISPRS Journal of Photogrammetry and Remote Sensing}, 167:\penalty0 190--200, 2020.

\bibitem[Thoker and Gall(2019)]{multi-modal_kd4}
Fida~Mohammad Thoker and Juergen Gall.
\newblock Cross-modal knowledge distillation for action recognition.
\newblock \emph{2019 IEEE International Conference on Image Processing (ICIP)}, pages 6--10, 2019.

\bibitem[Tian et~al.(2024)Tian, Aggarwal, Colaco, Kira, and Gonzalez-Franco]{diffuse_attend_segment_zeroshot_seg_diff}
Junjiao Tian, Lavisha Aggarwal, Andrea Colaco, Zsolt Kira, and Mar Gonzalez-Franco.
\newblock Diffuse attend and segment: Unsupervised zero-shot segmentation using stable diffusion.
\newblock In \emph{Proceedings of the IEEE/CVF Conference on Computer Vision and Pattern Recognition}, pages 3554--3563, 2024.

\bibitem[Toker et~al.(2024)Toker, Eisenberger, Cremers, and Leal-Taix{\'e}]{satsynth}
Aysim Toker, Marvin Eisenberger, Daniel Cremers, and Laura Leal-Taix{\'e}.
\newblock Satsynth: Augmenting image-mask pairs through diffusion models for aerial semantic segmentation.
\newblock In \emph{Proceedings of the IEEE/CVF Conference on Computer Vision and Pattern Recognition}, pages 27695--27705, 2024.

\bibitem[Trabucco et~al.(2023)Trabucco, Doherty, Gurinas, and Salakhutdinov]{effectivedataaugmentationdiffusion}
Brandon Trabucco, Kyle Doherty, Max Gurinas, and Ruslan Salakhutdinov.
\newblock Effective data augmentation with diffusion models.
\newblock \emph{arXiv preprint arXiv:2302.07944}, 2023.

\bibitem[Tung and Mori(2019)]{sp}
Frederick Tung and Greg Mori.
\newblock Similarity-preserving knowledge distillation.
\newblock In \emph{Proceedings of the IEEE/CVF International Conference on Computer Vision (ICCV)}, pages 1365--1374, 2019.

\bibitem[Vespe and Greidanus(2012)]{intro4_darkspot}
M. Vespe and H. Greidanus.
\newblock Sar image quality assessment and indicators for vessel and oil spill detection.
\newblock \emph{IEEE Transactions on Geoscience and Remote Sensing}, 50\penalty0 (11):\penalty0 4726--4734, 2012.

\bibitem[Wang et~al.(2022)Wang, Lohit, Jones, and Fu]{what_KD}
Huan Wang, Suhas Lohit, Michael~N Jones, and Yun Fu.
\newblock What makes a "good" data augmentation in knowledge distillation - a statistical perspective.
\newblock In \emph{Advances in Neural Information Processing Systems}, pages 13456--13469. Curran Associates, Inc., 2022.

\bibitem[Wang et~al.(2019)Wang, Sheng, Liu, Chen, Wan, and Mao]{intro10_speckle_noise}
W. Wang, H. Sheng, S. Liu, Y. Chen, J. Wan, and J. Mao.
\newblock An edge-preserving active contour model with bilateral filter based on hyperspectral image spectral information for oil spill segmentation.
\newblock In \emph{Proceedings of the Workshop on Hyperspectral Image and Signal Processing: Evolution in Remote Sensing (WHISPERS)}, pages 1--5, 2019.

\bibitem[Wang et~al.(2020)Wang, Zhou, Jiang, Bai, and Xu]{wang2020intra}
Yukang Wang, Wei Zhou, Tao Jiang, Xiang Bai, and Yongchao Xu.
\newblock Intra-class feature variation distillation for semantic segmentation.
\newblock In \emph{Computer Vision--ECCV 2020: 16th European Conference, Glasgow, UK, August 23--28, 2020, Proceedings, Part VII 16}, pages 346--362. Springer, 2020.

\bibitem[Wei et~al.(2024)Wei, Luo, and Luo]{SKD}
Shicai Wei, Chunbo Luo, and Yang Luo.
\newblock Scaled decoupled distillation.
\newblock In \emph{Proceedings of the IEEE/CVF Conference on Computer Vision and Pattern Recognition (CVPR)}, pages 15975--15983, 2024.

\bibitem[Wu et~al.(2023)Wu, Zhao, Shou, Zhou, and Shen]{wu2023diffumask}
Weijia Wu, Yuzhong Zhao, Mike~Zheng Shou, Hong Zhou, and Chunhua Shen.
\newblock Diffumask: Synthesizing images with pixel-level annotations for semantic segmentation using diffusion models.
\newblock In \emph{Proceedings of the IEEE/CVF International Conference on Computer Vision}, pages 1206--1217, 2023.

\bibitem[Wu et~al.(2024)Wu, Wong, Yu, Shi, Kwok, and Zou]{SAM_OIL}
Wenhui Wu, Man~Sing Wong, Xinyu Yu, Guoqiang Shi, Coco Yin~Tung Kwok, and Kang Zou.
\newblock Compositional oil spill detection based on object detector and adapted segment anything model from sar images.
\newblock \emph{IEEE Geoscience and Remote Sensing Letters}, 2024.

\bibitem[Xie et~al.(2021)Xie, Wang, Yu, Anandkumar, Alvarez, and Luo]{experiment_segformer}
Enze Xie, Wenhai Wang, Zhiding Yu, Anima Anandkumar, Jose~M Alvarez, and Ping Luo.
\newblock Segformer: Simple and efficient design for semantic segmentation with transformers.
\newblock \emph{Advances in neural information processing systems}, 34:\penalty0 12077--12090, 2021.

\bibitem[Ye et~al.(2020)Ye, Ji, Wang, Gao, and Song]{ye2020data-free-kd}
Jun Ye, Yang Ji, Xiaoyu Wang, Xin Gao, and Mingli Song.
\newblock Data-free knowledge amalgamation via group-stack dual-gan.
\newblock In \emph{Proceedings of the IEEE/CVF Conference on Computer Vision and Pattern Recognition (CVPR)}, pages 14745--14754, 2020.

\bibitem[Yu et~al.(2023)Yu, Li, Lou, Liu, Wan, Chen, and Li]{intro19_da_label_nuclei_image_seg}
Xinyi Yu, Guanbin Li, Wei Lou, Siqi Liu, Xiang Wan, Yan Chen, and Haofeng Li.
\newblock Diffusion-based data augmentation for nuclei image segmentation.
\newblock In \emph{International Conference on Medical Image Computing and Computer-Assisted Intervention}, pages 592--602. Springer, 2023.

\bibitem[Zagoruyko and Komodakis(2017)]{AT}
Sergey Zagoruyko and Nikos Komodakis.
\newblock Paying more attention to attention: Improving the performance of convolutional neural networks via attention transfer.
\newblock In \emph{International Conference on Learning Representations}, 2017.

\bibitem[Zhang et~al.(2019)Zhang, Zhu, and Ye]{Zhang_2019_CVPR}
Feng Zhang, Xiatian Zhu, and Mao Ye.
\newblock Fast human pose estimation.
\newblock In \emph{Proceedings of the IEEE/CVF conference on computer vision and pattern recognition}, pages 3517--3526, 2019.

\bibitem[Zhang et~al.(2020)Zhang, Feng, Luo, Li, Wei, and Li]{related_oilspill_quad}
Jin Zhang, Hao Feng, Qingli Luo, Yu Li, Jujie Wei, and Jian Li.
\newblock Oil spill detection in quad-polarimetric sar images using an advanced convolutional neural network based on superpixel model.
\newblock \emph{Remote Sensing}, 12\penalty0 (6), 2020.

\bibitem[Zhang et~al.(2024)Zhang, Li, Lu, Fang, McKinnon, Tsin, Quan, and Yao]{jointnet}
Jingyang Zhang, Shiwei Li, Yuanxun Lu, Tian Fang, David McKinnon, Yanghai Tsin, Long Quan, and Yao Yao.
\newblock Jointnet: Extending text-to-image diffusion for dense distribution modeling.
\newblock \emph{International Conference on Learning Representations (ICLR)}, 2024.

\bibitem[Zhang et~al.(2023)Zhang, Wu, Ren, Li, Qin, Xiao, Liu, Wang, Zheng, and Ma]{diffusionengine}
Manlin Zhang, Jie Wu, Yuxi Ren, Ming Li, Jie Qin, Xuefeng Xiao, Wei Liu, Rui Wang, Min Zheng, and Andy~J Ma.
\newblock Diffusionengine: Diffusion model is scalable data engine for object detection.
\newblock \emph{arXiv preprint arXiv:2309.03893}, 2023.

\bibitem[Zhao et~al.(2022)Zhao, Cui, Song, Qiu, and Liang]{Zhao_2022_CVPR}
Borui Zhao, Quan Cui, Renjie Song, Yiyu Qiu, and Jiajun Liang.
\newblock Decoupled knowledge distillation.
\newblock In \emph{Proceedings of the IEEE/CVF Conference on Computer Vision and Pattern Recognition (CVPR)}, pages 11953--11962, 2022.

\bibitem[Zhao et~al.(2018)Zhao, Li, Alsheikh, Tian, Zhao, Torralba, and Katabi]{multi-modal_kd2}
Mingmin Zhao, Tianhong Li, Mohammad~Abu Alsheikh, Yonglong Tian, Hang Zhao, Antonio Torralba, and Dina Katabi.
\newblock Through-wall human pose estimation using radio signals.
\newblock In \emph{2018 IEEE/CVF Conference on Computer Vision and Pattern Recognition}, pages 7356--7365, 2018.

\bibitem[Zhu et~al.(2020)Zhu, Li, Zhang, and Guan]{related_remotesegmentation3}
Qiqi Zhu, Zhen Li, Yanan Zhang, and Qingfeng Guan.
\newblock Building extraction from high spatial resolution remote sensing images via multiscale-aware and segmentation-prior conditional random fields.
\newblock \emph{Remote Sensing}, 12\penalty0 (23):\penalty0 3983, 2020.

\bibitem[Zhu et~al.(2021{\natexlab{a}})Zhu, Deng, Zheng, Zhong, Guan, Lin, Zhang, and Li]{related_remotesegmentation2}
Qiqi Zhu, Weihuan Deng, Zhuo Zheng, Yanfei Zhong, Qingfeng Guan, Weihua Lin, Liangpei Zhang, and Deren Li.
\newblock A spectral-spatial-dependent global learning framework for insufficient and imbalanced hyperspectral image classification.
\newblock \emph{IEEE Transactions on Cybernetics}, 52\penalty0 (11):\penalty0 11709--11723, 2021{\natexlab{a}}.

\bibitem[Zhu et~al.(2021{\natexlab{b}})Zhu, Zhang, Wang, Zhong, Guan, Lu, Zhang, and Li]{related_remotesegmentation1}
Qiqi Zhu, Yanan Zhang, Lizeng Wang, Yanfei Zhong, Qingfeng Guan, Xiaoyan Lu, Liangpei Zhang, and Deren Li.
\newblock A global context-aware and batch-independent network for road extraction from vhr satellite imagery.
\newblock \emph{ISPRS Journal of Photogrammetry and Remote Sensing}, 175:\penalty0 353--365, 2021{\natexlab{b}}.

\bibitem[Zhu et~al.(2022)Zhu, Zhang, Li, Yan, Guan, Zhong, Zhang, and Li]{cbdnet}
Qiqi Zhu, Yanan Zhang, Ziqi Li, Xiaorui Yan, Qingfeng Guan, Yanfei Zhong, Liangpei Zhang, and Deren Li.
\newblock Oil spill contextual and boundary-supervised detection network based on marine sar images.
\newblock \emph{IEEE Transactions on Geoscience and Remote Sensing}, 60:\penalty0 1--10, 2022.

\end{thebibliography}
}

\clearpage
\setcounter{page}{1}
\maketitlesupplementary
\appendix

\section{Introduction}
\noindent In this supplementary material, we present a comprehensive analysis of our diffusion-based Data Augmentation and Knowledge Transfer (DAKTer) strategy for the data scarcity problem in SAR oil spill segmentation.
First, we provide implementation details for training with our DAKTer strategy.
Next, we describe a detailed derivation process of our SNR-based balancing factor \(b\).
We also discuss a limitation of our DAKTer strategy.
Lastly, we present additional experimental results including qualitative results of generated SAR images and segmentation masks from our DAKTer strategy and comparisons of segmentation performance.

\section{Implementation Details}
\label{imp_details}

\subsection{Training a diffusion-based DA model.}
For generation of realistic SAR oil spill images and their soft labels, we employ a pixel-domain diffusion model, DDPM \cite{ddpm}.
For stable training, we adopt a two-step training approach: the DDPM is first trained using bilinearly down-sampled image-mask pairs at a low resolution (\(128\times128\)), and then we fine-tune at a high resolution (\(256\times256\)).
We train for \(300k, 100k\) steps for the low resolution with batch size 64 and \(100k, 50k\) steps for the high resolution with batch size 16 on the OSD \cite{related_oilspill_compare_all} and SOS dataset \cite{cbdnet}, respectively.
We use AdamW optimizer \cite{adamW} with a learning rate of \(2e^{-5}\) in the first stage, and \(1e^{-5}\) for the next stage.
We utilize a linear noise scheduler with \(1,000\) steps for training, while we use the DDIM \cite{song2020denoising} scheduler with 200 steps for inference.
Training is conducted on a Nvidia A6000 GPU.
In the OSD dataset \cite{related_oilspill_compare_all}, we cropped images into non-overlapping $256 \times 256$ patches, except for the final section, where we applied overlapping cropping to ensure full coverage of the image. Additionally, we used only 20\% of the images in which all pixels belonged to the 'sea-surface' class.

\subsection{Knowledge transfer to student models.}
To evaluate the effectiveness of our data augmentation and knowledge transfer with soft labels, we generate an augmented dataset with the same number of samples as the original training image-mask pairs. 
Using the augmented dataset and the original dataset \cite{related_oilspill_compare_all, cbdnet}, we train four recent segmentation models \cite{exp_deeplabv3+, cbdnet, cheng2021mask2former, experiment_segformer} for 400 epochs with a learning rate of \(1e^{-4}\) and a batch size of 32.
Specifically, we employ backbones of Swin-Tiny \cite{cheng2021mask2former} and SegFormer-B3 \cite{experiment_segformer} for Mask2Former \cite{cheng2021mask2former} and SegFormer \cite{experiment_segformer} for all experiments.
The hyperparameter $T$ is set to 2 for \cref{eq:soft_label} and the loss weights $\lambda_{\text{ce}}$, $\lambda_{\text{kd}}$, and $\lambda_{\text{dice}}$ are empirically set to $1$, $0.1$, and $0.5$, respectively.

\begin{table*}
\scriptsize
\centering
\setlength{\tabcolsep}{5pt} 
\begin{tabular}{cccccccccccc}
\toprule
 & & & & & & & \multicolumn{5}{c}{Per-class IoU (\%)} \\
 \cmidrule(lr){8-12} 
 Seg. Network & DA Methods & mIoU (\%) & F1 (\%) & Precision (\%) & Recall (\%) & Accuracy (\%) &  Sea Surface  & Oil Spill & Look-alike & Ship & Land \\ 
\midrule
\multirow{5}{*}{DeepLabV3+ \cite{exp_deeplabv3+}} & - & 65.10 & 76.46 & 78.54 & 75.88 & 97.73 & 94.15 & 46.62 & 41.70 & 49.48 & 93.55 \\
    & SemGAN \cite{semgan} & 66.79 & 78.21 & 80.54 & 77.09 & 96.94 & 91.77 & 45.80 & 49.62 & 53.28 & 93.49 \\
  & DDPM \cite{ddpm} & 68.61 & 79.77 & 81.11 & 78.97 & 97.26 & 92.57 & 51.58 & 54.24 & 51.17 & 93.46 \\
  & SatSynth \cite{satsynth} & 68.24 & 79.36 & 81.23 & 78.17 & 97.23 & 92.45 & 51.49 & 53.30 & 49.13 & 94.83 \\
  & DAKTer (Ours) & 69.35 & 80.24 & 80.40 & 80.73 & 97.37 & 92.79 & \textbf{53.49} & \textbf{56.40} & 48.73 & 95.32 \\
\midrule
\multirow{5}{*}{CBD-Net \cite{cbdnet}} & - & 66.32 & 77.60 & 79.19 & 76.83 & 98.14 & 95.29 & 47.86 & 45.68 & 50.16 & 92.58 \\
    & SemGAN \cite{semgan} & 67.76 & 79.11 & 78.66 & 80.12 & 97.10 & 92.01 & 51.51 & 53.36 & 49.03 & 92.87 \\
  & DDPM \cite{ddpm} & 68.58 & 79.68 & 79.73 & 80.37 & 97.09 & 92.11 & 49.16 & 52.40 & 54.44 & 94.78 \\
  & SatSynth \cite{satsynth} & 68.98 & 80.13 & 82.18 & 78.46 & 97.34 & 92.84 & 50.65 & 53.75 & 55.18 & 92.46 \\
  & DAKTer (Ours) & 70.49 & 81.24 & 81.29 & 82.01 & 97.23 & 92.37 & \textbf{54.29} & \textbf{55.87} & 54.18 & 95.73 \\
\midrule
\multirow{5}{*}{Mask2Former \cite{cheng2021mask2former}} & - & 65.95 & 77.30 & 79.83 & 75.38 & 98.15 & 95.44 & 48.25 & 47.22 & 46.69 & 92.14 \\
    & SemGAN \cite{semgan} & 66.16 & 77.45 & 82.46 & 73.69 & 97.39 & 93.01 & 47.85 & 51.31 & 43.93 & 94.68 \\
    & DDPM \cite{ddpm} & 66.84 & 78.02 & 79.40 & 76.99 & 97.41 & 93.04 & 49.32 & 53.60 & 43.41 & 94.81 \\
    & SatSynth \cite{satsynth} & 66.04 & 77.22 & 77.60 & 77.32 & 97.16 & 92.25 & \textbf{50.83} & 52.83 & 38.78 & 95.52 \\
  & DAKTer (Ours) & 68.22 & 79.33 & 80.04 & 79.19 & 97.51 & 93.35 & 50.40 & \textbf{58.03} & 46.09 & 93.26 \\
\midrule
\multirow{5}{*}{SegFormer \cite{experiment_segformer}} & - & 67.46 & 78.48 & 78.32 & 78.76 & 98.24 & 95.52 & 48.92 & 46.79 & 51.61 & 94.47\\
    & SemGAN \cite{semgan} & 67.66 & 79.04 & 80.56 & 77.77 & 97.43 & 93.01 & 51.80 & 54.17 & 47.88 & 91.45 \\
    & DDPM \cite{ddpm} & 69.77 & 80.61 & 80.90 & 80.48 & 97.57 & 93.43 & 51.30 & 57.33 & 51.93 & 94.87 \\
    & SatSynth \cite{satsynth} & 69.79 & 80.60 & 80.97 & 80.80 & 97.35 & 92.66 & 54.83 & 55.88 & 49.77 & 95.79 \\
  & DAKTer (Ours) & 70.74 & 81.35 & 81.64 & 81.49 & 97.82 & 94.06 & \textbf{56.47} & \textbf{60.12} & 48.55 & 94.48 \\
\bottomrule
\end{tabular}
\caption{Comparison of segmentation performance on the OSD dataset \cite{related_oilspill_compare_all} using recent segmentation models \cite{exp_deeplabv3+, cbdnet, cheng2021mask2former, experiment_segformer} trained with original dataset and 100\% augmented datasets synthesized by SemGAN \cite{semgan}, DDPM \cite{ddpm}, SatSynth \cite{satsynth}, and our DAKTer.}
\label{tab:supp_multiclass_iou}
\end{table*}

\section{Detailed Derivation of the Balancing Factor}
\label{sec:detailed_derivation_fo_b}
In \cref{sec:balancing_joint_generataion}, we introduce an SNR-based balancing factor \(b\) to stabilize the joint generation of SAR images and segmentation masks.
To quantify the amount of noise corruption to each modality, we utilize signal-to-noise ratio (SNR), calculated via the 2D Discrete Fourier Transform (DFT).
For image $\mathbf{x} \in \mathbb{R}^{H \times W}$, the 2D DFT is defined as:
\begin{equation}
    \label{eq:f_x}
    F_\mathbf{x}(u, v) = \sum_{h=0}^{H-1} \sum_{w=0}^{W-1} \mathbf{x}(h, w) \cdot e^{-i2\pi \left( \frac{uh}{H} + \frac{vw}{W} \right)},
\end{equation}
where $F_\mathbf{x}(u, v)$ represents the complex frequency component of $\mathbf{x}$ at the spatial frequency $(u,v)$. 
Each complex frequency component of a scaled image $a \cdot \mathbf{x}$ can be expressed as:
\begin{equation}
    \begin{aligned}\
    \label{eq:f_ax}
    F_{a\mathbf{x}}(u, v) = a \cdot F_{\mathbf{x}}(u, v).
\end{aligned}
\end{equation}
The mean power of image $\mathbf{x}$ in frequency domain is defined as: 
\begin{equation}
    \text{P}(\mathbf{x}) = \mathbb{E} [F_\mathbf{x}(u, v) \cdot F_\mathbf{x}^*(u, v)],
\end{equation}
where \(*\) denotes the complex conjugate. 
For the scaled image $a \cdot \mathbf{x}$, its mean power is given as:
\begin{equation}
\begin{aligned}
    \text{P}(a\mathbf{x}) &= \mathbb{E} [F_{a\mathbf{x}}(u, v) \cdot F_{a\mathbf{x}}^*(u, v)] \\
    &= \mathbb{E} [aF_\mathbf{x}(u, v) \cdot aF_\mathbf{x}^*(u, v)] \\
    &= a^2 \cdot \mathbb{E} [F_\mathbf{x}(u, v) \cdot F_\mathbf{x}^*(u, v)] \\
    &= a^2 \cdot \text{P}(\mathbf{x}).
\end{aligned}
\end{equation}
Using this property, we derive our SNR-based balancing factor \(b\). In the forward diffusion process, the noise-corrupted SAR image $\mathbf{x}_t$ and its corresponding segmentation mask $\mathbf{y}_t$ at timestep $t$ are expressed as follows:
\begin{equation}
    \begin{aligned}
        \mathbf{x}_t = \alpha_t \mathbf{x}_0+\sigma_t \bm{\epsilon},
        \quad \mathbf{y}_t = \alpha_t \mathbf{y}_0+ \sigma_t \bm{\epsilon}.
    \end{aligned}
\end{equation}
where $\alpha_t$ and $\sigma_t$ are scalars determined by the noise scheduler, representing the degree of signal and noise at timestep $t$, respectively. 
The SNR is defined as the ratio of the mean power of the signal to the mean power of the noise. So, the SNRs for $\mathbf{x}_t$ and $\mathbf{y}_t$ are given by:
\begin{equation}
    \begin{aligned}
        \text{SNR}(\mathbf{x}_t) = \frac{\alpha_t^2\text{P}(\mathbf{x}_0)}{\sigma_t^2\text{P}(\bm{\epsilon})}, \quad
        \text{SNR}(\mathbf{y}_t) = \frac{\alpha_t^2\text{P}(\mathbf{y}_0)}{\sigma_t^2\text{P}(\bm{\epsilon})}. 
    \end{aligned}
\end{equation}
To balance the noise corruption process between a SAR image and its corresponding mask, we scale the original segmentation mask $\mathbf{y}_0$ by the SNR-baed balancing factor \(b\), resulting in a scaled noise-corrupted segmentation mask \(\mathbf{y}'_t\), which is given by:
\begin{equation}
    \begin{aligned}    
        \quad \mathbf{y}'_t &= \alpha_t b \mathbf{y}_0+ \sigma_t \bm{\epsilon}. \\
    \end{aligned}
\end{equation}
The SNR for $\mathbf{y}'_t$ is then given by:
\begin{equation}
    \begin{aligned}
        \text{SNR}(\mathbf{y}'_t) 
        &= \frac{\text{P}(\alpha_t b \mathbf{y}_0)}{\text{P}(\sigma_t\bm{\epsilon})} \\
        &= \frac{\alpha_t^2 b^2  \text{P}(\mathbf{y}_0)}{\sigma_t^2\text{P}(\bm{\epsilon})} \\
        &= b^2 \cdot \text{SNR}(\mathbf{y}_t).
    \end{aligned}
\end{equation}
The ratio of the retained information levels between \(\mathbf{x}_t\) and \(\mathbf{y}'_t\) is expressed as follows:
\begin{equation}
    \begin{aligned}
        \frac{\text{SNR}(\mathbf{x}_t)}{{\text{SNR}(\mathbf{y}'_t)}} &=
        \frac{\text{SNR}(\mathbf{x}_t)}{{b^2 \cdot \text{SNR}(\mathbf{y}_t)}}.
    \end{aligned}
\end{equation}
To make this ratio equal to 1, we obtain our SNR-based balancing factor $b$ as follows:
\begin{equation}
    \begin{aligned}
    \label{eq:balancing_factor_suppl}
        b &= \sqrt{\frac{\text{SNR}(\mathbf{x}_t)}{{\text{SNR}(\mathbf{y}_t)}}} \\
        &= \sqrt{\frac{\text{P}(\mathbf{x}_0)}{{\text{P}(\mathbf{y}_0)}}}.
    \end{aligned}
\end{equation}
Our SNR-based balancing factor ensures maintaining balanced information levels of different modalities between SAR images and masks in the forward diffusion process, which is critical for learning joint distribution.

\section{Limitations}
While our DAKTer strategy successfully generates SAR images and their corresponding soft labels using DDPM \cite{ddpm}, it is currently constrained to a resolution of $256 \times 256$.
Extending DDPM, a pixel-domain diffusion model, to generate spatially larger images, such as $512 \times 512$ or $1024 \times 1024$, requires significantly more computational resources.
Recently, the latent diffusion model (LDM) \cite{rombach2022high} has demonstrated successful generation capability for high-resolution natural images by operating a diffusion process in the latent domain, which is typically reduced to $1/8$ or $1/4$ of the original image size.
However, we observed that the LDM struggles to synthesize realistic SAR images with speckle noise in the latent domain, which is typically reduced to $1/8$ or $1/4$ of the original image size.
Future work should aim to develop efficient models capable of generating high-resolution SAR images while preserving their unique noise characteristics.

\section{Additional Experimental Results}

\subsection{Qualitative Comparison of Data Generation}
\begin{figure*}[!t]
    \centering
    \includegraphics[width=\linewidth]{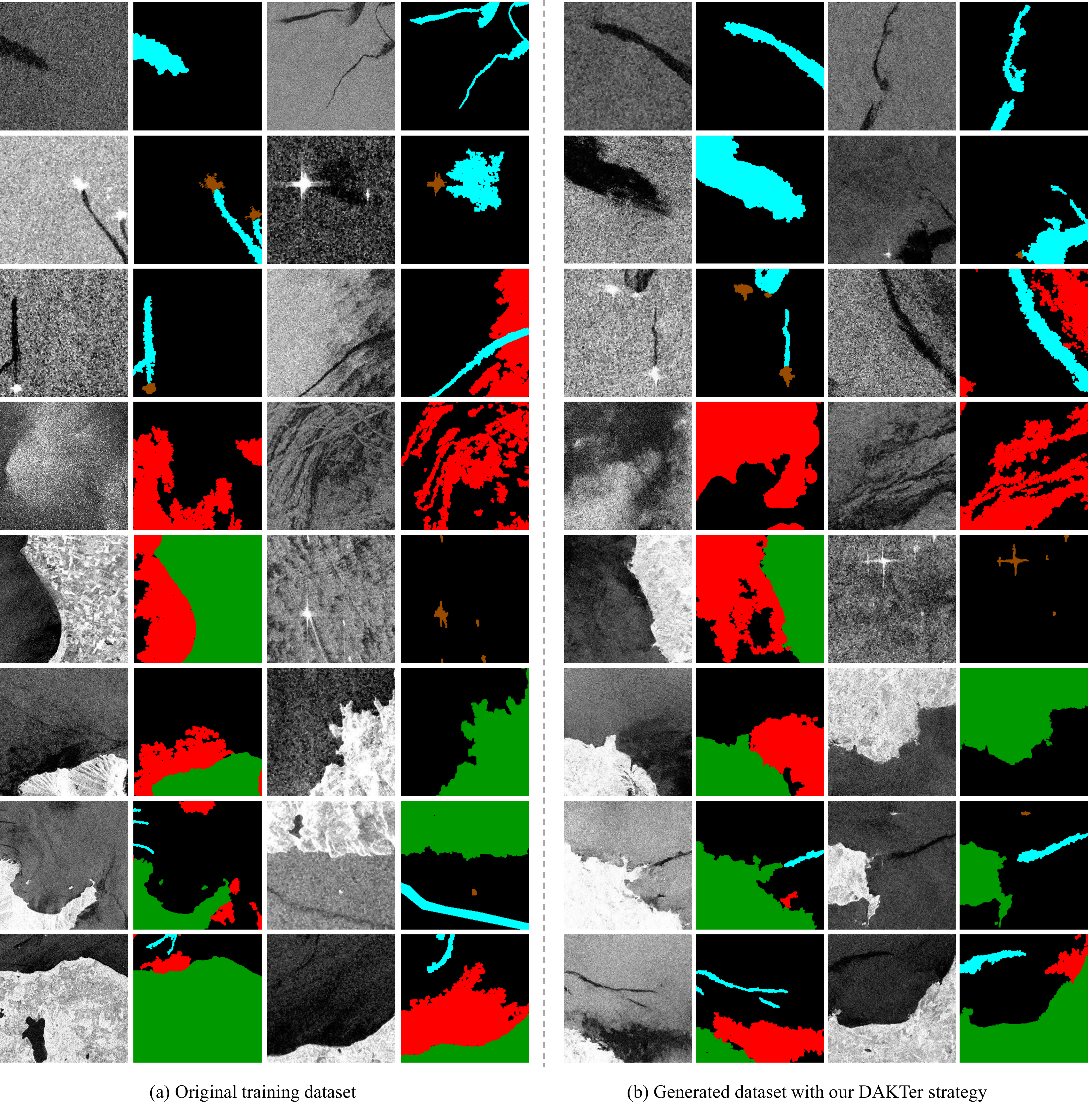}
    \caption{Qualitative results of generated SAR images and their corresponding generated segmentation masks. (a) shows several original training samples (SAR images and their associated segmentation masks) of the OSD dataset \cite{related_oilspill_compare_all}, (b) shows several generated samples from DDPM with our DAKTer strategy.}  
    \label{fig:sppl_ql_osd}  
\end{figure*}

\begin{figure*}[!t]
    \centering
    \includegraphics[width=\linewidth]{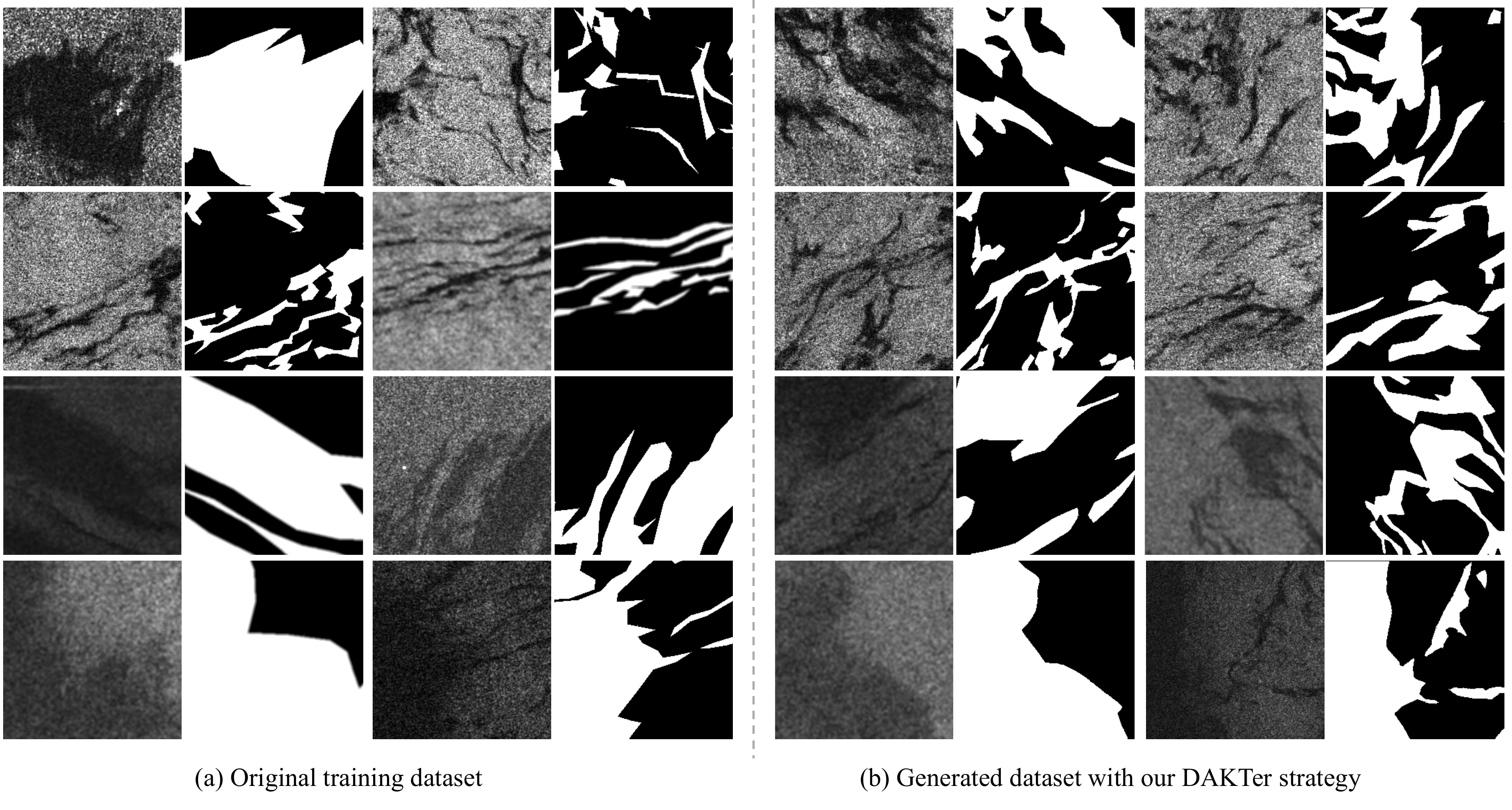}
    \caption{Qualitative results of generated SAR images and their corresponding generated segmentation masks. (a) shows several original training samples (SAR images and their associated segmentation masks) of the SOS-Sentinel dataset \cite{cbdnet}, (b) shows several generated samples from DDPM with our DAKTer strategy.}  
    \label{fig:sppl_ql_sentinel}  
\end{figure*}

\begin{figure*}[!t]
    \centering
    \includegraphics[width=\linewidth]{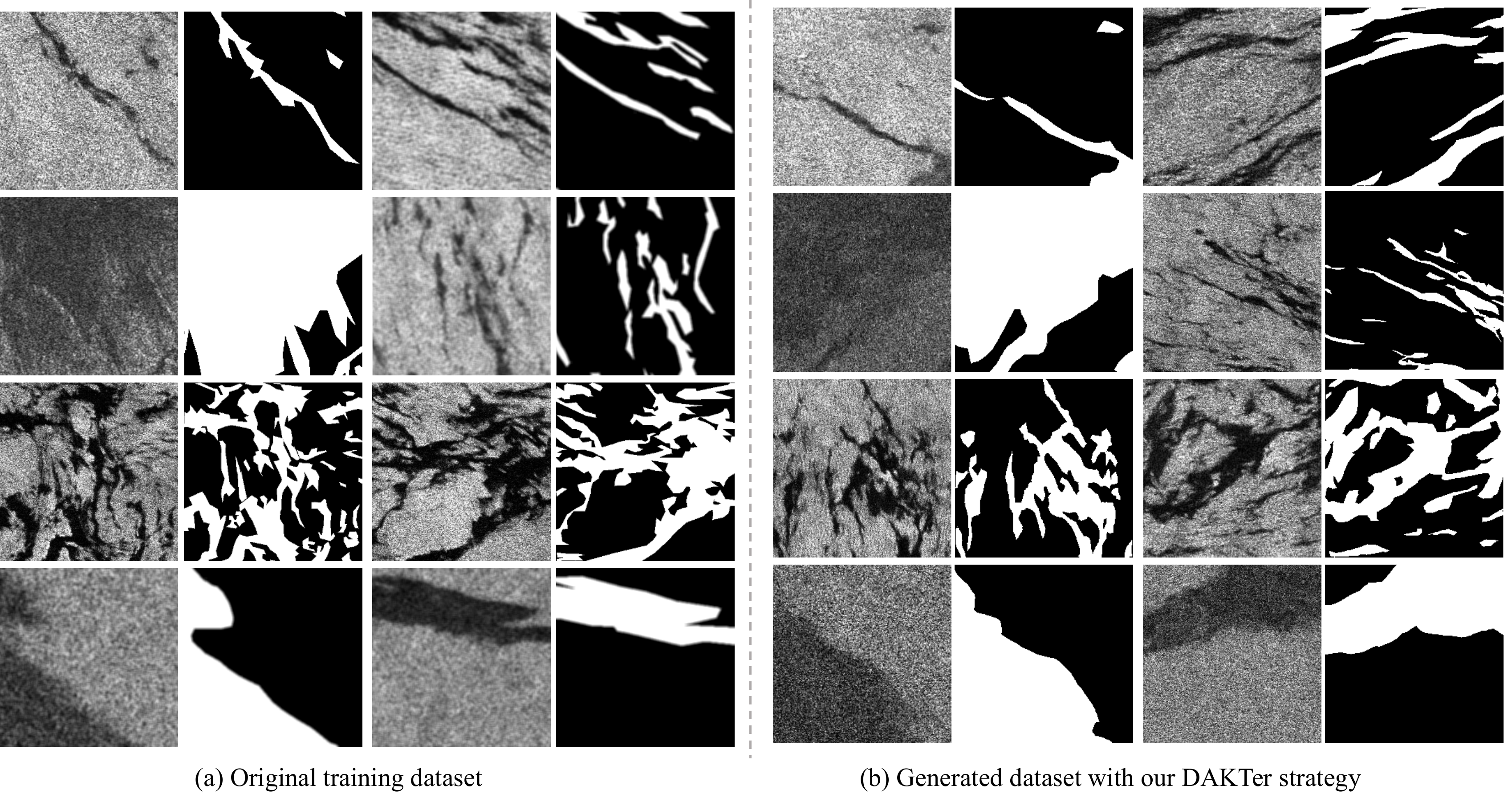}
    \caption{Qualitative results of generated SAR images and their corresponding generated segmentation masks. (a) shows several original training samples (SAR images and their associated segmentation masks) of the SOS-ALOS dataset \cite{cbdnet}, (b) shows several generated samples from DDPM with our DAKTer strategy.}  
    \label{fig:sppl_ql_palsar}  
\end{figure*}

We provide additional generated pairs of SAR images and their segmentation masks from DDPM trained with our DAKTer strategy, comparing to the original training dataset (OSD dataset \cite{related_oilspill_compare_all}) in \cref{fig:sppl_ql_osd}. 
Furthermore, \cref{fig:sppl_ql_sentinel} and \cref{fig:sppl_ql_palsar} show SAR images and segmentation masks from the original training dataset of SOS-Sentinel and SOS-ALOS \cite{cbdnet}, as well as the respective pairs generated by DDPM trained with ours.
The results demonstrate that DDPM trained with DAKTer strategy generates realistic SAR images and segmentation masks with high correspondences, effectively capturing the characteristics of the original datasets.

\subsection{Segmentation Performance}

In \cref{tab:supp_multiclass_iou}, we evaluate the effectiveness of our DAKTer strategy on the recent segmentation models \cite{exp_deeplabv3+, cbdnet, cheng2021mask2former, experiment_segformer} using all segmentation metrics and per-class IoU on the OSD dataset \cite{related_oilspill_compare_all}.
As shown, our DAKTer strategy consistently improves the segmentation performance with large margins.
Also, it should be noted that segmentation models, trained with our DAKTer strategy, are significantly improved on `oil-spill' and `look-alike' classes (highlighted in bold) compared to other DA methods \cite{semgan, ddpm, satsynth}.
This superior performance can be attributed to our DAKTer strategy leveraging per-pixel class probabilities as knowledge. By incorporating this information, our approach effectively distinguishes between 'oil-spill' and 'look-alike' regions, which often exhibit similar patterns in SAR oil spill images.

\begin{figure*}[t!]
    \centering
    \includegraphics[width=0.8\linewidth]{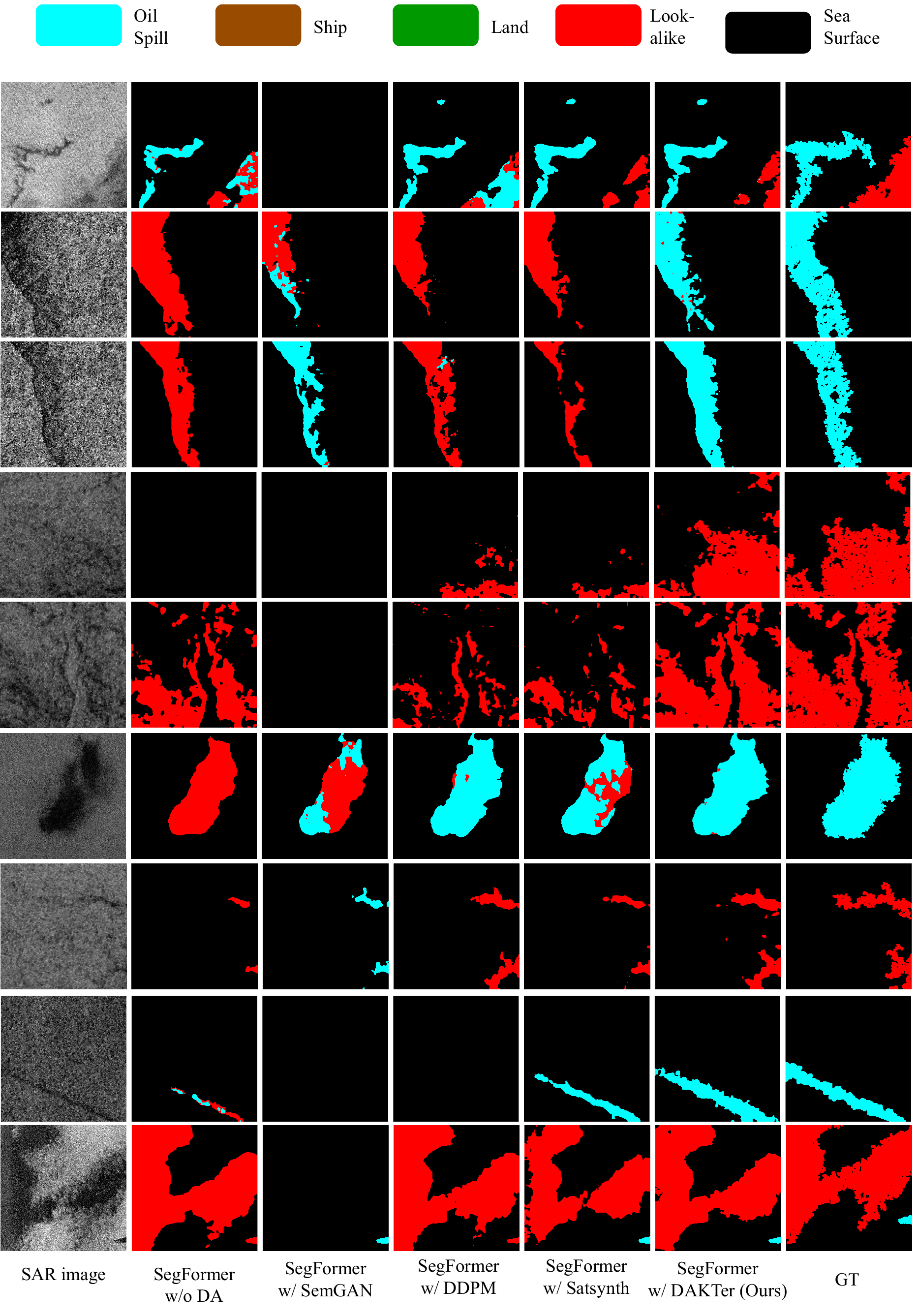}
    \caption{Qualitative comparison of segmentation results between DA methods (without DA, SemGAN \cite{semgan}, DDPM \cite{ddpm}, SatSynth \cite{satsynth}, and our DAKTer strategy) evaluated on the OSD dataset \cite{related_oilspill_compare_all}. All experiment is conducted by training SegFormer \cite{experiment_segformer} but using different augmented datasets from different DA methods along with the original dataset.}
    \label{fig:DAKTer_osd_segformer}
\end{figure*}

In \cref{fig:DAKTer_osd_segformer}, we compare segmentation results from SegFormer \cite{experiment_segformer} trained with different data augmentation methods \cite{semgan, ddpm, satsynth} and our DAKTer strategy against the baseline model trained with the original dataset.
In SAR images, `oil spill' and `look-alike' regions both appear as dark areas, making them difficult to distinguish using traditional segmentation models. However, SegFormer trained with DAKTer effectively captures subtle differences between these regions by leveraging the rich class probability distributions provided by soft labels. This reduces false positive in look-alike areas and enhances the overall segmentation accuracy of oil spill segmentation.
These results demonstrate that our DAKTer strategy transfers rich knowledge through soft-label-based supervision to student segmentation models.

\begin{figure*}[!t]
    \centering
    \includegraphics[width=0.85\linewidth]{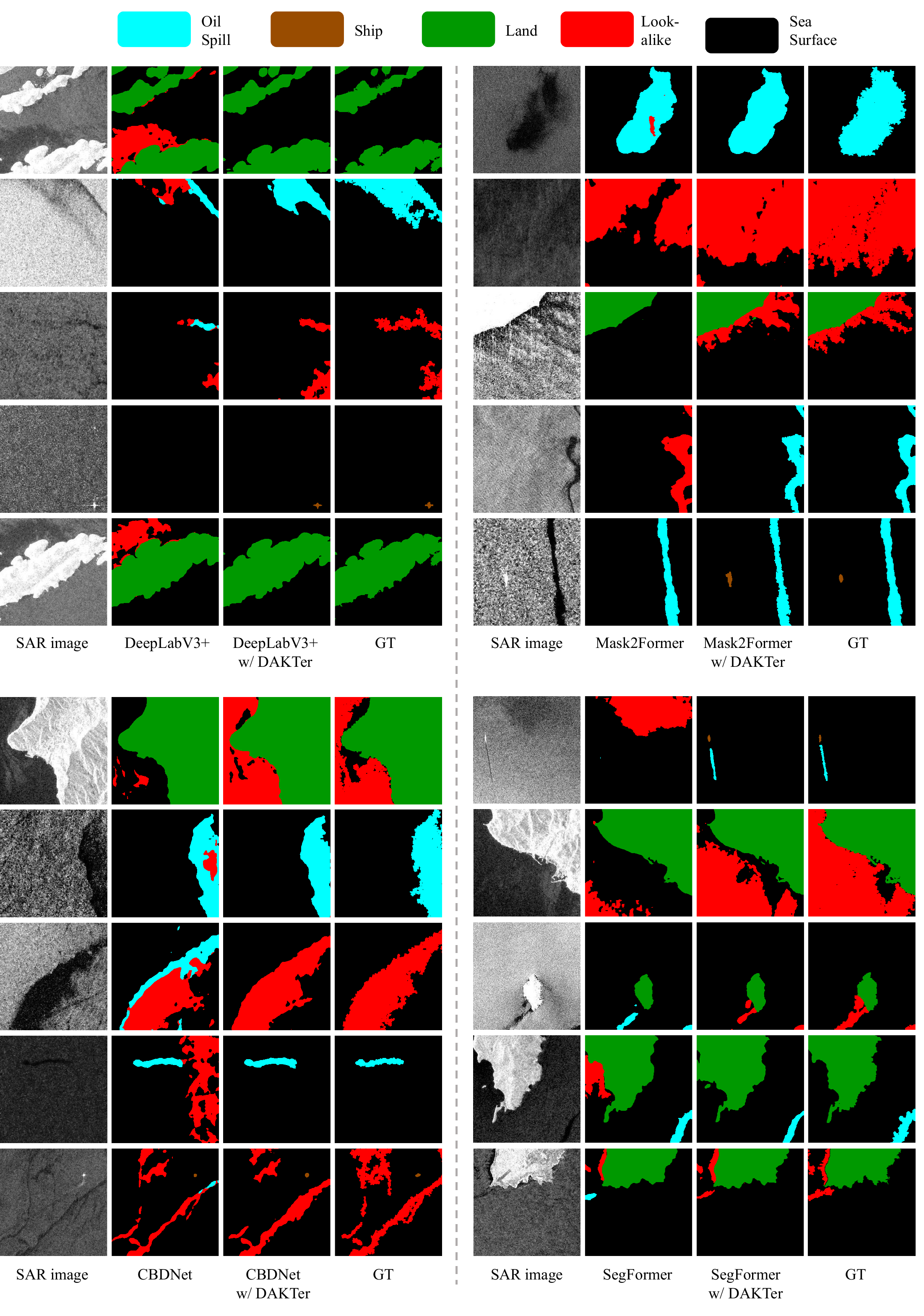}
    \caption{Qualitative comparison of segmentation results between segmentation models \cite{exp_deeplabv3+, cheng2021mask2former, cbdnet, experiment_segformer} trained on the OSD dataset \cite{related_oilspill_compare_all} without DA model against the models trained with the augmented dataset from our DAKTer strategy.}  
    \label{fig:supp_osd}  
\end{figure*}

\begin{figure*}[!t]
    \centering
    \includegraphics[width=0.85\linewidth]{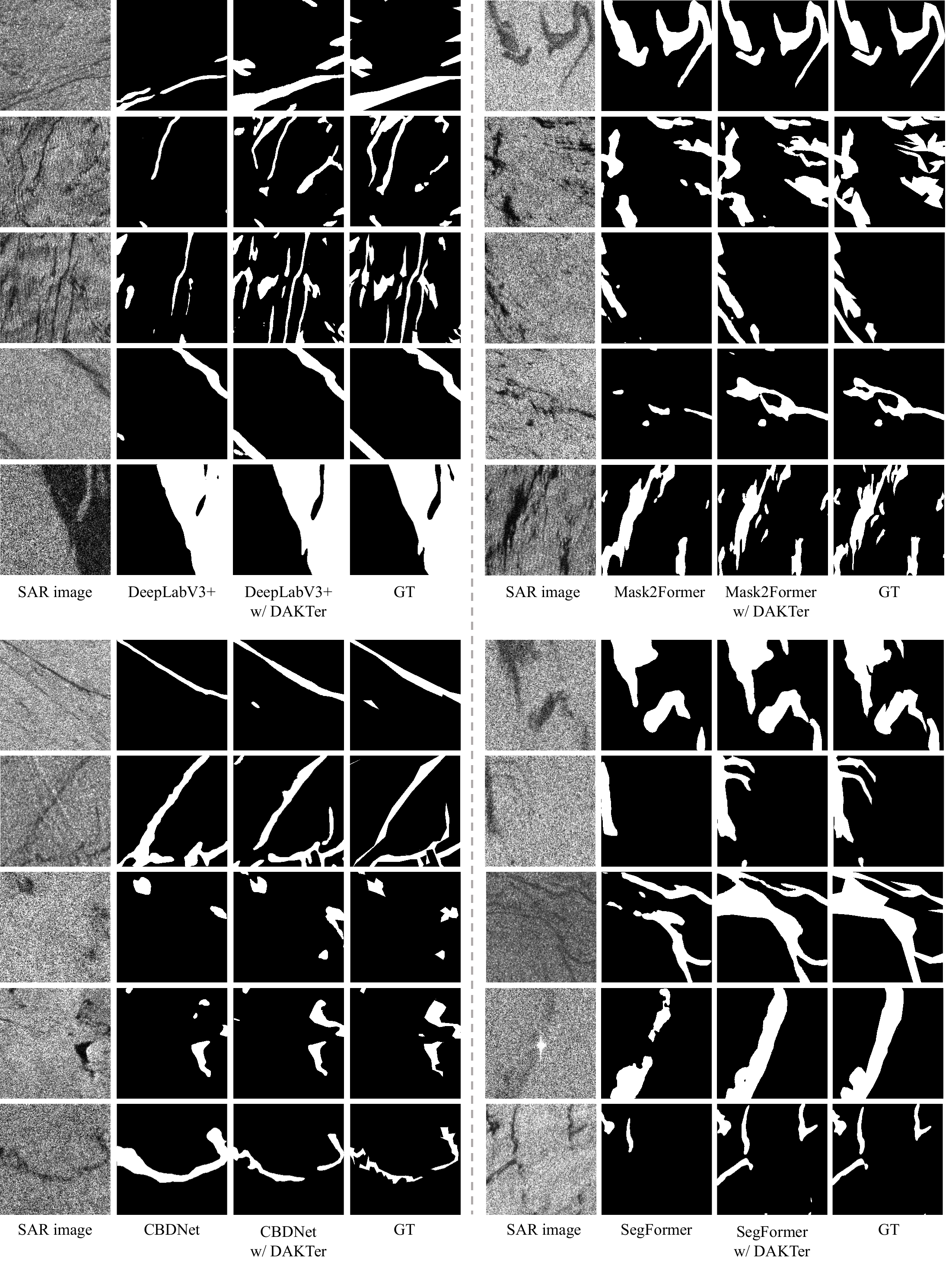}
    \caption{Qualitative comparison of segmentation results between segmentation models \cite{exp_deeplabv3+, cheng2021mask2former, cbdnet, experiment_segformer} trained on the SOS-ALOS dataset \cite{cbdnet} without DA model against the models trained with the augmented dataset from our DAKTer strategy.}  
    \label{fig:supp_ALOS}  
\end{figure*}

\begin{figure*}[!t]
    \centering
    \includegraphics[width=0.85\linewidth]{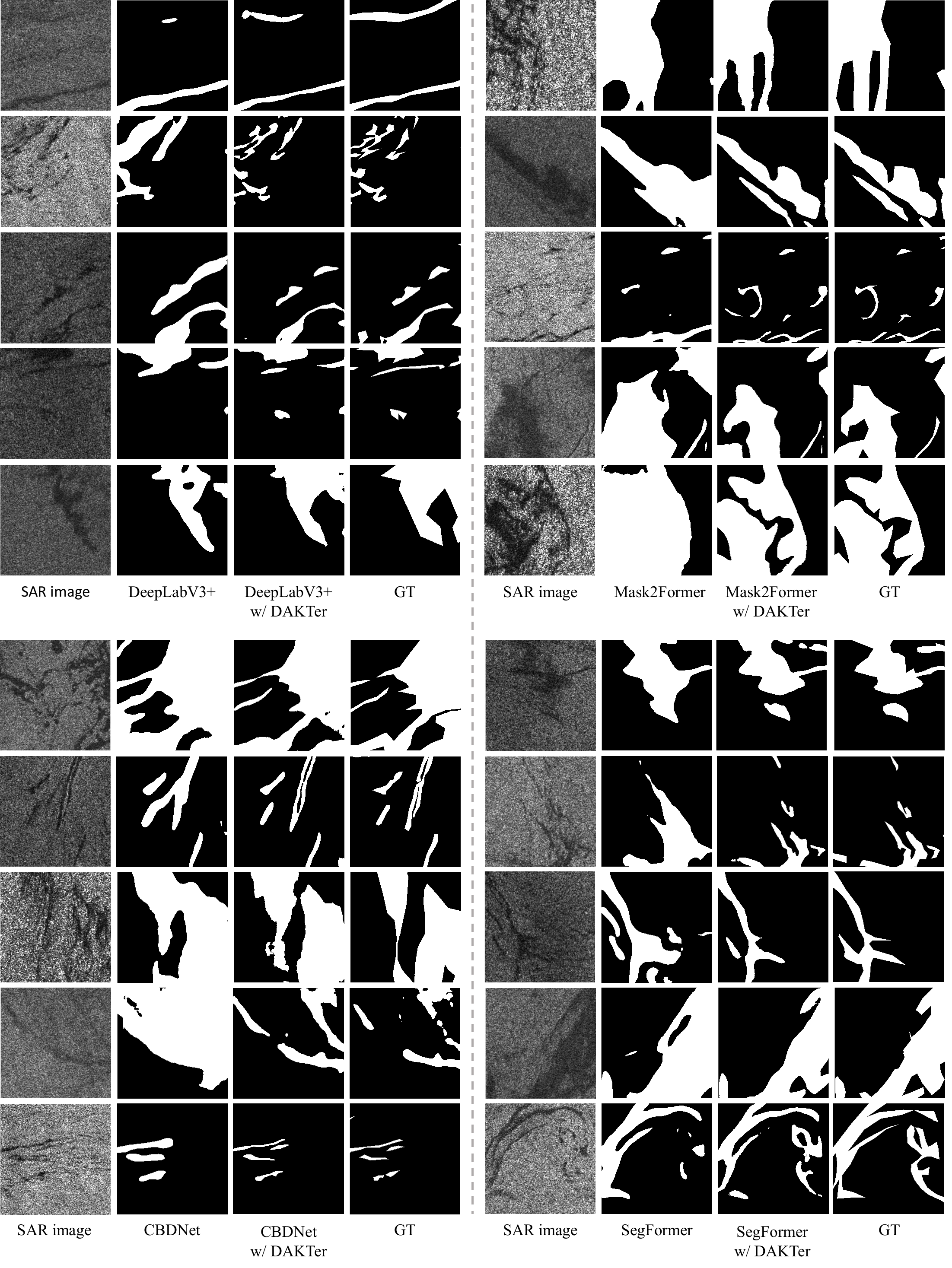}
    \caption{Qualitative comparison of segmentation results between segmentation models \cite{exp_deeplabv3+, cheng2021mask2former, cbdnet, experiment_segformer} trained on the SOS-Sentinel dataset \cite{cbdnet} without DA model against the models trained with the augmented dataset from our DAKTer strategy.}  
    \label{fig:supp_sentinel}  
\end{figure*}

\cref{fig:supp_osd}, \cref{fig:supp_ALOS}, and \cref{fig:supp_sentinel} present qualitative results on the OSD dataset \cite{related_oilspill_compare_all}, SOS-ALOS, and SOS-Sentinel dataset \cite{cbdnet}, respectively.
As shown, our DAKTer strategy effectively boosts segmentation models \cite{exp_deeplabv3+, cbdnet, cheng2021mask2former, experiment_segformer} by generating augmented datasets and transferring knowledge with generated soft labels.
Notably, thanks to our knowledge transfer from generated soft labels, the segmentation models \cite{exp_deeplabv3+, cbdnet, cheng2021mask2former, experiment_segformer} trained with our DAKTer strategy effectively distinguish oil spill regions from look-alike regions. 
These results emphasize the robustness and effectiveness of our DAKTer strategy in addressing data scarcity problem in SAR oil spill segmentation tasks.



\clearpage

\end{document}